\documentclass{article}

\usepackage[nonatbib, preprint]{neurips_2021}
\usepackage[utf8]{inputenc} 
\usepackage[T1]{fontenc}    
\usepackage[colorlinks=true,
linkcolor=red,
urlcolor=blue,
citecolor=blue]{hyperref}
\usepackage{url}            
\usepackage{booktabs}       
\usepackage{amsfonts}       
\usepackage{nicefrac}       
\usepackage{microtype}      
\usepackage{amsthm}
\usepackage{upgreek}
\usepackage{bbm}

\usepackage{mdframed}
\newmdenv[
    rightline=true,
    bottomline=false,
    topline=false,
    leftline=false,
    rightmargin = 0pt,
    innertopmargin=0pt,
    innerbottommargin=0pt,
    innerrightmargin=0pt,
    innerleftmargin=0pt,
    linewidth=0.75pt]{rline}
\newmdenv[
    rightline=true,
    bottomline=false,
    topline=false,
    leftline=false,
    rightmargin = 0pt,
    innertopmargin=0pt,
    innerbottommargin=0pt,
    innerrightmargin=0pt,
    innerleftmargin=0pt,
    linewidth=0.75pt]{rline2}

\newtheorem{lemma}{Lemma}

\newtheorem{Prop}{Proposition}
\newtheorem{theorem}{Theorem}

\newtheorem{Corollary}{Corollary}

\newtheorem{Exa}{Example}
\newtheorem*{Exa*}{Example}

\usepackage[numbers, compress]{natbib}
\makeatletter
\def\set@curr@file#1{%
  \begingroup
    \escapechar\m@ne
    \xdef\@curr@file{\expandafter\string\csname #1\endcsname}%
  \endgroup
}

\def\quote@name#1{"\quote@@name#1\@gobble""}
\def\quote@@name#1"{#1\quote@@name}
\def\unquote@name#1{\quote@@name#1\@gobble"}
\makeatother
\usepackage{graphics}

\graphicspath{{./img/}}

\usepackage{amsmath, amssymb, amsfonts}
\usepackage{enumitem}
\usepackage{comment}
\usepackage{dsfont}
\usepackage{xargs}
\usepackage[textwidth=3cm,textsize=footnotesize]{todonotes}

\newtheorem{assum}{A\hspace{-2pt}}

\usepackage{subcaption}
\usepackage{cleveref}
\usepackage{hyperref}

\crefname{assum}{A\hspace{-2pt}}{A\hspace{-2pt}}
\crefname{assumb}{B\hspace{-2pt}}{B\hspace{-2pt}}
\crefname{assumUGE}{UGE\hspace{-1pt}}{UGE\hspace{-1pt}}
\crefname{assumUE}{UE\hspace{-1pt}}{UE\hspace{-1pt}}
\crefname{assumSUP}{M\hspace{-1pt}}{M\hspace{-1pt}}
\crefname{assumR}{R\hspace{-1pt}}{R\hspace{-1pt}}

\crefname{lemma}{lemma}{lemmas}
\Crefname{lemma}{Lemma}{Lemmas}
\crefname{Exa}{example}{examples}
\Crefname{Exa}{Example}{Examples}
\crefname{Prop}{proposition}{propositions}
\Crefname{Prop}{Proposition}{Propositions}
\crefname{Exa}{example}{examples}
\Crefname{Exa}{Example}{Examples}
\crefname{Corollary}{corollary}{corollaries}
\Crefname{Corollary}{Corollary}{Corollaries}

\newcommand{\PE}{\mathbb{E}}
\newcommand{\PP}{\mathbb{P}}

\def\rset{\mathbb{R}}

\def\nset{\ensuremath{\mathbb{N}}}
\def\nsets{\ensuremath{\mathbb{N}^*}}
\def\nsetm{\ensuremath{\mathbb{N}_-}}
\def\zset{\ensuremath{\mathbb{Z}}}

\newcommand{\SG}{\operatorname{SG}}

\newcommand{\mrl}{\mathrm{L}}

\newcommand{\bConst}[1]{\operatorname{C}_{{#1}}}

\def\sfA{\mathsf{A}}
\def\sfB{\mathsf{B}}
\def\sfC{\mathsf{C}}
\def\sfD{\mathsf{D}}

\newcommandx\sequence[3][2=,3=]
{\ifthenelse{\equal{#3}{}}{\ensuremath{\{ #1_{#2}\}}}{\ensuremath{\{ #1_{#2}\!~:\!~#2 \in #3 \}}}}
\newcommandx\sequenceD[2][2=]
{\ifthenelse{\equal{#2}{}}{\ensuremath{\{ #1\}}}{\ensuremath{\{ #1\!~:\!~#2  \}}}}
\newcommandx\sequenceDouble[4][3=,4=]
{\ifthenelse{\equal{#3}{}}{\ensuremath{\{ (#1_{#3},#2_{#3})\}}}{\ensuremath{\{ (#1_{#3},#2_{#3})\!~:\!~#3 \in #4 \}}}}

\newcommandx\sequencen[2][2=n]
{\ensuremath{\{ #1_{#2}\!~:\!~#2\in\nset\}}}
\newcommandx\sequencens[2][2=n]
{\ensuremath{\{ #1_{#2} \!~:\!~#2\in\nsets\}}}
\newcommandx\sequencet[4]
{\ensuremath{\{ #1{#2_{#3}} \, : \, \eqsp #3 \in #4 \}}}
\def\PE{\mathbb{E}}
\def\P{\mathbb{P}}
\def\ProdB{\Gamma}

\def\Ga{G^{(\alpha)}}
\def\ProdBa{\ProdB^{(\alpha)}}

\newcommandx{\PVar}[1][1=]{\ensuremath{\operatorname{Var}_{#1}}}

\def\Sigmabf{\boldsymbol{\Sigma}}
\def\lineG{\Sigmabf}
\def\lineGa{\Sigmabf^{\alpha}}
\newcommand{\abs}[1]{\vert #1\vert}

\newcommandx{\norm}[2][2=]{\Vert#1 \Vert_{{#2}}}
\newcommandx{\normLigne}[2][2=]{\Vert#1 \Vert_{{#2}}}
\newcommandx{\normop}[2][2=]{\Vert{#1}\Vert_{{#2}}}
\newcommandx{\normopLigne}[2][2=]{\Vert{#1}\Vert_{{#2}}}
\newcommandx{\normopLine}[2][2=]{\Vert{#1}\Vert_{{#2}}}
\newcommandx{\osc}[2][1=]{\mathrm{osc}_{#1}(#2)}

\newcommand{\iid}{i.i.d.}

\newcommandx{\as}[1][1=\PP]{\ensuremath{#1\, -\mathrm{a.s.}}}

\newcommand{\ie}{i.e.}

\newcommand{\eqsp}{\;}

\newcommand{\Id}{\mathrm{I}}

\def\ttheta{\tilde{\theta}}
\def\utheta{\tilde{\theta}^{\sf (tr)}}
\def\vtheta{\tilde{\theta}^{\sf (fl)}}

\newcommand{\ConstD}{\mathsf{D}}

\newcommand{\Jnalpha}[2]{J_{#1}^{(\alpha,#2)}}
\newcommand{\Hnalpha}[2]{H_{#1}^{(\alpha,#2)}}
\newcommand{\Jnalphawn}[1]{J^{(\alpha,#1)}}
\newcommand{\Hnalphawn}[1]{H^{(\alpha,#1)}}

\newcommand{\cointLigne}[1]{[#1)}
\newcommand{\ocint}[1]{\left(#1\right]}
\newcommand{\ocintLigne}[1]{(#1]}
\newcommand{\ooint}[1]{\left(#1\right)}
\newcommand{\oointLigne}[1]{(#1)}
\newcommand{\ccint}[1]{\left[#1\right]}

\newcommandx{\Nnorm}[2][1=V]{[ #2]_{#1}}

\newcommandx{\CPE}[3][1=]{{\mathbb E}^{#3}_{#1}\left[#2 \right]}

\def\thetalim{\theta^\star}

\newcommand{\rme}{\mathrm{e}}
\newcommand{\rmd}{\mathrm{d}}

\def\funnoisew{\varepsilon}

\def\funcctilde{\tilde{c}_u}

\newcommandx{\funcct}[2][1=]{\funcctilde^{#1}(#2)}

\def\qcond{\kappa_{Q}}
\def\tqcond{\kappa_{\tQ}}
\def\qcondtild{\kappa_{\mathsf{\tilde{Q}}}}

\newcommand{\1}{\mathbbm{1}}

\def\plusinfty{+\infty}

\DeclareMathAlphabet{\mathpzc}{OT1}{pzc}{m}{it}

\newcommand{\txts}{\textstyle}

\newcommand{\proba}[1]{\mathbb{P}\left( #1 \right)}

\newcommandx\probaMarkovTilde[2][2=]
{\ifthenelse{\equal{#2}{}}{{\widetilde{\mathbb{P}}_{#1}}}{\widetilde{\mathbb{P}}_{#1}\left[ #2\right]}}

\newcommand{\expe}[1]{\PE \left[ #1 \right]}

\newcommand{\parenthese}[1]{\left(#1 \right)}
\newcommand{\parentheseLigne}[1]{(#1 )}
\newcommand{\parentheseDeux}[1]{\left[ #1 \right]}
\newcommand{\parentheseDeuxLigne}[1]{[ #1 ]}
\newcommand{\defEns}[1]{\left\lbrace #1 \right\rbrace }

\def\half{\nicefrac{1}{2}}

\def\bA{\bar{A}}
\def\tA{\tilde{A}}
\def\tbfA{\mathbf{\tA}}
\def\ta{\tilde{a}}

\def\X{{\bf X}}
\def\Y{{\bf Y}}
\def\balpha{\bar{\alpha}}

\def\thetas{\thetalim}
\def\sphere{\mathbb{S}}
\def\realpart{\mathrm{Re}}
\def\transpose{\top}
\def\tQ{\tilde{Q}}
\def\Am{{\bf A}}
\def\bm{{\bf b}}

\def\barb{\bar{b}}
\def\KLs{\mathrm{KL}}
\def\thetalima{\theta_{\infty}^{(\alpha)}}
\def\tthetalima{\tilde{\theta}_{\infty}^{(\alpha)}}

\def\Pens{\mathcal{P}}
\def\mrl{\mathrm{L}}
\def\rml{\mathrm{L}}

\def\Cov{\mathrm{Cov}}
\def\transpose{\top}
\def\mcb{\mathcal{B}}
\def\minfty{-\infty}
\newcommand{\convproba}[1]{\overset{\PP}{\underset{#1}{\longrightarrow}}}
\def\barA{\bar{A}}

\def\Zbf{\mathbf{Z}}
\def\Abf{\mathbf{A}}
\def\Bbf{\mathbf{B}}

\def\bfA{\Abf}
\def\bfB{\Bbf}
\def\eps{\varepsilon}
\def\talpha{\tilde{\alpha}}
\def\tvarphi{\tilde{\varphi}}

\title{Tight High Probability Bounds for Linear Stochastic Approximation with Fixed Stepsize}

\author{%
  Alain Durmus
    \thanks{Authors listed in alphabetical order.
    } 
    \\
  ENS Paris-Saclay \\
  \texttt{alain.durmus@ens-paris-saclay.fr} \\
  \And
  Eric Moulines \\
  Ecole Polytechnique \\
  and HSE University \\
  \texttt{eric.moulines@polytechnique.edu}\\
   \And
   Alexey Naumov  \\
  HSE University\\
  \texttt{anaumov@hse.ru} \\
  \And Sergey Samsonov \\
  HSE University\\
  \texttt{svsamsonov@hse.ru} \\
  \And Kevin Scaman \\
  INRIA, DI/ENS, PSL Research University\\
  \texttt{kevin.scaman@gmail.com} \\
  \And Hoi-To Wai \\
  The Chinese University of Hong Kong\\
  \texttt{htwai@se.cuhk.edu.hk} 
 }
\date{February 2021}

\begin{document}

\maketitle

\begin{abstract}
This paper provides a non-asymptotic analysis of linear stochastic approximation (LSA) algorithms with fixed stepsize. This family of methods arises in many machine learning tasks and is used to obtain approximate solutions of a linear system $\bar{A}\theta = \bar{b}$ for which $\bar{A}$ and $\bar{b}$ can only be accessed through random estimates $\{({\bf A}_n, {\bf b}_n): n \in \mathbb{N}^*\}$.  Our analysis is based on new results regarding moments and high probability bounds for products of matrices which are shown to be tight. We derive high probability bounds on the performance of LSA under weaker conditions on the sequence $\{({\bf A}_n, {\bf b}_n): n \in \mathbb{N}^*\}$ than previous works. However, in contrast, we establish polynomial concentration bounds with order depending on the stepsize. We show that our conclusions cannot be improved  without additional assumptions on the sequence of random matrices $\{{\bf A}_n: n \in \mathbb{N}^*\}$, and in particular that no Gaussian or exponential high probability bounds can hold.  Finally, we pay a particular attention to establishing  bounds with sharp order with respect to the number of iterations and the stepsize and  whose leading terms contain the covariance matrices appearing in the central limit theorems.
\end{abstract}

\section{Introduction} \label{sec:intro}

This paper provides a detailed analysis of Linear Stochastic Approximation (LSA) schemes which aim at finding a solution $\thetas$ for a linear system of the form $\bA \theta = \barb$. In particular, we analyze LSA with  a fixed stepsize $\alpha >0$ which consists in defining a sequence of estimates $\sequence{\theta}[n][\nset]$ for $\thetas$  by the recursion
\begin{equation} 
\label{eq:lsa}
\theta_{n+1} = \theta_n - \alpha \{ \Am_{n+1} \theta_n - \bm_{n+1} \} \eqsp, \quad n \in\nset\eqsp,
\end{equation}
where $\sequenceDouble{\Am}{\bm}[n][\nsets]$ is a sequence of \iid~random
variables used as proxy for $\bA \in \rset^{d \times d }$ and
$\barb \in \rset^d$ which are typically unknown.  This class of
algorithms and the corresponding setting have a long history and 
important applications in signal processing such as channel
equalization and echo cancellation
\cite{benveniste2012adaptive,kushner2003stochastic}. It has renewed
interests in machine learning and computational statistics especially
for least-square estimation, Reinforcement learning (RL)
and $Q$-learning
\cite{bertsekas2003parallel,bottou2018optimization,watkins1992q,sutton:td:1988}.
The recursion \eqref{eq:lsa} has already been studied in depth in
several works which derive asymptotic
\cite{polyak1992acceleration,kushner2003stochastic,borkar:sa:2008,benveniste2012adaptive}
and non-asymptotic
\cite{rakhlin2012making,nemirovski2009robust,bach:moulines:2013,jain2018accelerating,pmlr-v99-jain19a,bhandari2018finite,lakshminarayanan:2018a,srikant:1tsbounds:2019,chen2020explicit, durmus2021stability} guarantees.  

However, in most cases, there is a consistent gap
between these two types of analyses. While asymptotic analysis gives 
important insights on the qualitative convergence of
\eqref{eq:lsa} based on statistical key quantities of the problem on
hand, they do not provide finite-time convergence, or high probability
bounds, necessary to obtain non-asymptotic confidence sets, see
\cite{nesterov2008confidence,chen2020statistical} and the references therein. On the other
hand, non-asymptotic studies are in general too coarse and lose
significant statistical information in their derivation. Further,
their upper bounds are generally loose when used in predicting the
actual performance of LSA.  We aim at filling this gap and provide
conditions on $\sequenceDouble{\Am}{\bm}[n][\nsets]$ ensuring tight
high probability bounds on the sequence
$\sequence{\theta}[n][\nset]$.

This problem has been addressed in several contributions but at the expense of strong conditions on the sequence $\sequenceDouble{\Am}{\bm}[n][\nsets]$.
 \citep{frikha2012concentration} provided concentration bounds for non-linear stochastic algorithms under a log-Sobolev condition which turns out to be hard to verify for most applications except for
 the Euler-Maruyama discretization scheme applied to 
 Stochastic Differential Equation.  \citep{pepin2021concentration} derived concentration inequalities but assuming that the innovations in \eqref{eq:lsa} are uniformly bounded.
In contrast, we aim at giving simple and mild conditions ensuring high probability bounds. 
More precisely, one of our key contributions (\Cref{th:HP_error_bound_iid}) is to show that under mild conditions on the sequence $\sequenceDouble{\Am}{\bm}[n][\nsets]$, for any $\delta \in (0,1)$, $n \in \nset$ and $u \in \sphere^{d-1}$, 
\begin{equation} \label{eq:whp}
\begin{split}
    & \mathbb{P} \left( | u^\top ( \theta_{n} - \theta^\star) | \leq 
    c \{ \sqrt{\alpha u^\top \lineG u} + \alpha \} \sqrt{\log(1/\delta)} + c \{ \rho_\alpha^n + \alpha p_0^2 \} \delta^{-\frac{1}{p_0}} 
    \right) \geq 1 - \delta \eqsp,
\end{split}
\end{equation}
where $\rho_\alpha \in (0,1)$, $c > 0$ is a constant independent of $n, \alpha, \delta$,  and $p_0 = o(\alpha^{-1/4})$. In the above, $\lineG$ is the unique solution of the Lyapunov  equation which naturally appears in central limit theorems for LSA with diminishing stepsize.  In addition, we show that the bound we get is tight with respect to $\alpha$ and $\delta$ in the case where we only assume that $-\PE[\Am_1] = -\bA$ is Hurwitz. Indeed, we provide counterexamples illustrating that for a fixed stepsize $\alpha$ and under the conditions that we consider, logarithmic dependence in $1/\delta$ cannot hold in \eqref{eq:whp} but only a polynomial one. Regarding the dependence with respect to $\alpha$, we extend \cite{pflug:1986} and show that for $\alpha$ small enough, $\sequence{\theta}[n][\nset]$ admits a unique stationary distribution $\pi_{\alpha}$ and establish a central limit theorem for this family of distribution as $\alpha \downarrow 0$ at rate $\sqrt{\alpha}$ and with asymptotic covariance matrix $\lineG$ appearing in \eqref{eq:whp}.

Finally, our proofs rely on a new analysis of product of matrices, extending the recent work of  \citep{huang2020matrix}. In particular, we establish conditions ensuring uniform bounds in $n$ of the $p$-th moments of $\mathbf{Y}_n \cdots \mathbf{Y}_1$, where $\sequence{\mathbf{Y}}[n][\nsets]$ is a sequence of independent matrices whose expected values have a spectral radius less than $1$. In comparison to existing results, the main challenge that we address here is that the  random matrices $\sequence{\Am}[n][\nsets]$ are not assumed to be almost surely symmetric.

The paper is organized as follows. Section~\ref{sec:lsa} formally discusses the assumptions on LSA for our analysis. Section~\ref{sec:matrix_concentration} presents the moment bound for product of random matrices. Using this result, Section~\ref{sec:finite-time-high} shows the high probability concentration inequality \eqref{eq:whp} and Section~\ref{sec:optim-deriv-bounds} shows the tightness of the bounds by deriving a central limit theorem for LSA.

\paragraph{Notations}
Denote $\nsets = \nset \setminus \{0\}$ and $\nsetm = \zset \setminus \nsets$.
Let $d \in \nsets$ and $Q$ be a symmetric positive definite $d \times d$ matrix. For $x \in \rset^d$, we denote $\norm{x}[Q]= \{x^\top Q x\}^{\half}$. For brevity, we set $\norm{x}= \norm{x}[\Id_d]$.  We denote $\normop{A}[Q]= \max_{\norm{x}[Q]=1} \norm{Ax}[Q]$, and the subscriptless norm $\normop{A} = \normop{A}[\Id]$ is the standard spectral norm. We denote the condition number of $Q$ as $\qcond = \lambda_{\sf min}^{-1}( Q )\lambda_{\sf max}( Q ) $.
We denote $\sphere^{d-1} = \{x \in \rset^{d} | \norm{x} = 1\}$. Let $A_{1},\ldots,A_N$ be $d$-dimensional matrices. We denote $\prod_{\ell=i}^j A_\ell = A_j \ldots A_i$ if $i\leq j$ and with the convention $\prod_{\ell=i}^j A_\ell = \Id_d$ if $i >j$. We say that a centered random variable (r.v.) $X$ is sub-Gaussian  with variance factor $\sigma^2$ and we denote $X \in \SG(\sigma^2)$ if for all $\lambda \in \rset$, $\log \PE[\rme^{\lambda X}] \leq \lambda^2\sigma^2/2$.
We define the Wasserstein distance of order $2$ between two probabilities measure $\mu$ and $\nu$ on $\rset^d$ as $W_2(\mu,\nu) = \inf_{\zeta \in \Pi(\mu,\nu)} \int_{\rset^{2d}} \norm{x-y}^2 \rmd \zeta(x,y)$, where $\Pi(\mu,\nu)$ is the set of probability measures on $(\rset^{2d},\mcb(\rset^{2d}))$ with marginals $\mu$ and $\nu$ respectively. Denote by $\Pens_2(\rset^d)$ the set of all probability measures on $\rset^d$ with the finite second moment.


\section{Linear Stochastic Approximation: Setting and Assumptions}
\label{sec:lsa}
Consider the LSA recursion \eqref{eq:lsa} with a deterministic initial point $\theta_0$. The main assumption required in this paper is as follows:
\begin{assum}
  \label{assum:noise}
 $\{(\Am_n, \bm_n)\}_{n\in\nsets}$ is an \iid~sequence satisfying the following conditions.
  \begin{enumerate}[wide, labelwidth=!, labelindent=0pt,label=(\roman*),itemsep=-.5ex, topsep=-.25ex]
\item \label{assum:b_subexp}
$\PE[\bm_1] = \barb$ and there exists  $\bConst{b}>0$ such that, for any $u \in \sphere^{d-1}$, $u^{\transpose} (\bm_1 - \barb) \in \SG(\bConst{b}^2)$.
  \item \label{assum:Abounded}
  There exists  $\bConst{A}>0$ such that
$\normop{\Am_1} \leq \bConst{A}$ almost surely.
  \item \label{assum:Hurwitzmatrices} The matrix $-\bA= -\PE[\Am_1]$ is Hurwitz, \ie~for any eigenvalue $\lambda$ of $\bA$, $\realpart(\lambda) >0$.
  \end{enumerate}
\end{assum}
Both conditions \Cref{assum:noise}-\ref{assum:b_subexp}, \ref{assum:Abounded} are standard in analysis of LSA, e.g., in \citep{dalal:td0:2017,srikant:1tsbounds:2019,macchi1983second}. 
Meanwhile, \Cref{assum:noise}-\ref{assum:Hurwitzmatrices} guarantees the existence of a unique solution $\thetas$ to $\bA \theta = \barb$. 
It is also a sufficient and necessary condition for the solution of the ordinary differential equation $\dot{\theta}_t = -\bA \theta_t$ to converge exponentially to $\thetas$ \cite[Lemma 4.1.2]{jacob:zwart:2012}. The same kind of result holds for the discrete system $\theta_{n+1}^{\sf d} - \theta_n^{\sf d} =  - \alpha \barA \theta_{n}^{\sf d}$. 
\begin{Prop}
\label{lem:Hurwitzstability}
Assume that $-\bA$ is a Hurwitz matrix. Then there exists a unique positive definite matrix $Q$ satisfying the Lyapunov equation 
$\bA^\top Q + Q \bA =  \Id$. In addition, setting 
\begin{equation}
\label{eq: kappa_def}
a = \normop{Q}^{-1}/2\eqsp, \quad 
\text{and} \quad \alpha_\infty = (1/2) \normop{\bA}[Q]^{-2} \normop{Q}^{-1} \eqsp,
\end{equation}
then for any $\alpha \in [0, \alpha_{\infty}]$,
we get $\normop{\Id - \alpha \bA}[Q]^2 \leq 1 - a \alpha$. 
If in addition $\alpha \le \normop{Q}^2$ then
$ 1 - a \alpha \geq 1/2$.
\end{Prop}
This result is well known but its proof can be found in \Cref{sec:proofs-crefl_hurwitz_proof} for completeness. The above proposition implies that the discrete system converges exponentially as $\| \theta_{n+1}^{\sf d} \| \leq \sqrt{\qcond} (1-a \alpha)^{n/2} \| \theta_0^{\sf d} \|$ for $\alpha \in (0, \alpha_\infty)$.

Recall that the aim of this paper is to derive high probability bounds on $u^\top \{\theta_n-\thetas\}$ for any $n \in \nset$, $u \in\sphere^{d-1}$. 
Below, we present a counterexample to show that under only \Cref{assum:noise}, if $\alpha >0$ is fixed, then there exists $\bar{p}>0$ such that $\lim_{n \to \plusinfty} \PE[\normLigne{\theta_n-\thetas}^p] = \plusinfty$ for $p \geq \bar{p}$. As a corollary, it is impossible to obtain exponential high probability bounds for $\sequenceD{\norm{\theta_n - \thetas}}[n \in \nset]$.\vspace{.1cm}

\begin{Exa}
  \label{exa:no_p_moments}
Consider \eqref{eq:lsa} with $d=1$ taking $\bm_n= 0$ for any $n\in\nsets$ and for
  $\sequencens{\Am}$ an \iid~sequence of \textit{biased} Rademacher r.v.s with parameter $q_{A} \in
  \ooint{1/2,1}$:
\begin{equation}
  \label{eq:particular_example}
  \Am_n =
  \begin{cases}
    1 & \text{ with probability $q_{A}$}\eqsp, \\
    -1 & \text{ with probability $1-q_{A}$}\eqsp.
  \end{cases}
\end{equation}
This choice is associated with $\thetas = 0$ and corresponds to the recursion: $  \theta_n = \prod_{k=1}^n (1-\alpha \Am_k) \theta_0$, for some $\theta_0 \neq 0$.
For any $p \geq 1$ and $\alpha \in\ooint{0,1}$, we have by definition,
    \begin{equation*}
    \PE\parentheseDeux{\abs{\theta_n}^p} = \{q_{A}(1-\alpha)^p+(1-q_{A})(1+\alpha)^p \}^n \abs{\theta_0}^p \eqsp.
  \end{equation*}
Using the lower bounds $(1-\alpha)^p \geq 1-\alpha p$ and $(1+\alpha)^p \geq 1+\alpha p  + p(p-1)\alpha^2/2$, we get for any $p \geq 1$ and $\alpha \in\ooint{0,1}$,
  \begin{equation*}
    \PE\parentheseDeux{\abs{\theta_n}^p} \geq \{1-p\alpha[(2q_{A}-1) -(p-1)\alpha(1-q_{A})/2]\}^n \abs{\theta_0}^p \eqsp.
  \end{equation*}
  If $\alpha \in \ooint{0,1}$ is fixed, then for any
  $p > \bar{p}_{q,\alpha} = 1+2(2q_{A}-1)/[\alpha(1-q_{A})]$,
  we have $\lim_{n\to \plusinfty} \PE\parentheseDeux{\abs{\theta_n}^p} =
  \plusinfty$. 
  On the other hand, if $\alpha \in \ooint{0, 2 (2q_{A}-1) / (1-q_A) }$, then
  $\lim_{n \to \plusinfty} \PE[\theta_n^2] =0$. 
  Therefore
  $\sequence{\theta}[n][\nset]$ converges in distribution to the Dirac
  measure at $0$ which corresponds to the unique stationary
  distribution of this sequence as a Markov chain. In such a
  case, this distribution admit $p$ moments for any $p \geq
  0$. 
  
  However, this result is specific to this particular case and
  does not hold if only \Cref{assum:noise} holds. 
  Consider $\sequence{\theta}[n][\nset]$
  defined by \eqref{eq:lsa} with $\sequence{\Am}[n][\nsets]$ given in
  \eqref{eq:particular_example} and $\sequence{\bm}[n][\nsets]$ be an
  \iid~sequence of zero-mean Gaussian random variables with unit
  variance independent of $\sequence{\Am}[n][\nsets]$. We show in \Cref{sec:proof-example_station_no_p_moments} that there
  exists $\alpha_{2,\infty}$ such that for any
  $\alpha \in \ocint{0,\alpha_{2,\infty}}$, the Markov chain
    $\sequence{\theta}[n][\nset]$ admits a unique
    invariant distribution $\pi_{\alpha}$ for any $\alpha >0$. Further, for
    any $\alpha \in \ocint{0,\alpha_{2,\infty}}$ there exists
      $p_\alpha \geq 1$ such that
      $\int_{\rset} \abs{\theta}^p \rmd \pi_{\alpha}(\theta) =
      \plusinfty$ for any $p \geq p_\alpha$.
\end{Exa}
It is, however, possible to obtain any $p$-th moment uniform bound for $\sequenceD{\norm{\theta_n - \thetas}}[n \in \nset]$ by strengthening \Cref{assum:noise}-\ref{assum:Hurwitzmatrices} to:
\begin{assum}
\label{ass:contraction} 
There exist $\ta \in \ooint{0,1}$, $\tilde\alpha_{\infty} >0$ and a definite positive $d$-dimensional matrix $\tQ$ such that almost surely, for any $\alpha \in \ocint{0,\tilde\alpha_{\infty}}$, $\norm{\Id-\alpha\Am_1}[\tQ] < 1-\ta \alpha$. 
\end{assum}
Examples such that \Cref{ass:contraction} holds include regularized linear regression, where we take $\Am_1 = \lambda \Id + {\bf a}_1 {\bf a}_1^\top$, for some $\lambda > 0$ and under the assumption that $\norm{{\bf a}_1}$ is bounded almost surely. The LSA recursion \eqref{eq:lsa} approximates the solution to $( \lambda \Id + \PE[ {\bf a}_1 {\bf a}_1^\top ] ) \theta = \barb$ which is guaranteed to have a unique solution.

On the other hand, examples where \Cref{ass:contraction} does not hold are common. For this, we consider TD(0) learning with linear function approximation. For a Markov Reward Process with ${\sf X}$ as the state space, ${\rm P}: {\sf X} \times {\cal X} \rightarrow [0,1]$ as the transition probability, ${\rm R} : {\sf X} \rightarrow \rset$ as the reward function, and $\gamma \in (0,1)$ as a discount factor, TD(0) learning is described as in \eqref{eq:lsa} with
\begin{equation}
    \Am_n = \phi( x_n ) \{ \phi( x_n ) - \gamma \phi( x_n' ) \}^\top, \quad \bm_n = {\rm R}(x_n) \phi(x_n) \eqsp,
\end{equation}
where $\phi: {\sf X} \rightarrow \rset^d$ is a feature map. A typical setting is when $x_n$ is drawn from the stationary distribution of ${\rm P}$ and $x_n' \sim {\rm P}(x_n, \cdot)$. It is easy to verify \Cref{assum:noise} provided that $\| \phi(x) \|$, ${\rm R}(x)$ are bounded for all $x \in {\sf X}$ \citep{tsitsiklis:td:1997}. However, \Cref{ass:contraction} is violated as $\Am_n$ is only rank-one.

Our next endeavor is to establish moment estimates on the product below:
\begin{equation}
    \label{eq:definition-Phi} \textstyle
\ProdBa_{m:n}  = \prod_{i=m}^n (\Id - \alpha \Am_i ) \eqsp, \quad m,n \in\nsets, \quad m \leq n \eqsp.
\end{equation}
We also define its expected value as $\Ga_{m:n} = \PE[ \ProdBa_{m:n} ] = (\Id - \alpha \bA )^{n-m+1}$. 

The above product naturally appears after re-centering the LSA recursion \eqref{eq:lsa}. For any $n \in \nsets$, 
\begin{equation}
\label{eq:LSA-recursion-main} 
\theta_{n} - \thetas = \bigl(\Id - \alpha\Am_{n}\bigr)\{\theta_{n-1} - \thetas\} + \alpha \funnoisew_{n} \eqsp,\quad \funnoisew_{n} = \bm_{n} - \barb - \{ \Am_{n} - \bA \} \thetas \eqsp.
\end{equation}
An easy induction implies that
\begin{equation} \textstyle
\label{eq:LSA_recursion_expanded} 
\theta_{n} - \thetas =  \utheta_{n} + \vtheta_n \eqsp, \quad \utheta_n =  \ProdBa_{1:n} \{ \theta_0 - \thetas \} \eqsp, \quad \vtheta_n =  \alpha \sum_{j=1}^n \ProdBa_{j+1:n} \funnoisew_j \eqsp.
\end{equation}
The decomposition \eqref{eq:LSA_recursion_expanded} highlights the two  sources of error in the estimation of $\thetas$ by $\sequence{\theta}[n][\nset]$ which will be separately tackled: $\sequence{\utheta}[n][\nset]$ corresponds to the transient (or bias) term and $\sequence{\vtheta}[n][\nset]$ to the fluctuation term. Both errors are controlled by the product of matrices $\ProdBa_{m:n}$, thereby motivating the study of the moment bound on $\ProdBa_{1:n}$ as we present next.

 \section{Moment and High-probability Bounds for Products of Random Matrices}
 \label{sec:matrix_concentration}
 
Recall from \Cref{lem:Hurwitzstability} that the expected value $\Ga_{1:n} = \PE[ \ProdBa_{1:n} ]$ decays exponentially with $n$, here we expect a similar phenomenon to hold for the moment bound of $\ProdBa_{1:n}$. 
Precisely, in this section, we show that if $p$ is fixed, then there exists $\alpha_p >0$ such that for any $\alpha \in \ocint{0,\alpha_p}$, the $p$-th moment of $\ProdBa_{m:n}$ decays exponentially with $n-m$.


To facilitate our discussions, we introduce the following notations.
For $B \in \rset^{d \times d}$, we denote by $( \sigma_\ell(B) )_{ \ell=1 }^d$ its singular values. For $p \geq 1$, the Schatten \(p\)-norm is denoted by $\norm{B}[p] = \{\sum_{\ell=1}^d \sigma_\ell^p(B)\}^{1/p}$. For $p, q \geq 1$ and random matrix $\X$, we write $\norm{\X}[p,q] = \{\PE [\norm{\X}[p]^q] \}^{1/q}$.

In the following, we present the main technical result on the product of general random matrices. The proof is based on the framework introduced in \cite{huang2020matrix}. 
\begin{Prop}
\label{th:general_expectation}
    Let $\sequence{\Y}[\ell][\nset]$ be an independent sequence and $P$ be a positive definite matrix. Assume that for each $\ell \in \nset$ there exist $m_\ell \in (0,1)$  and $\sigma_{\ell} > 0$ such that \(\norm{\PE[\Y_\ell]}[P]^2  \leq 1 - m_\ell\) and \(\norm{\Y_\ell - \PE[\Y_\ell]}[P] \leq \sigma_{\ell}\) almost surely.  Define $\Zbf_n = \prod_{\ell = 0}^n \Y_\ell= \Y_n \Zbf_{n-1}$, for $n \geq 1$ and starting from $\Zbf_0$. Then, for any $2 \le q \le p$ and $n \geq 1$,
    \begin{equation} \label{eq:gen_expectation}
        \norm{\Zbf_n}[p,q]^2 \leq \kappa_P \prod_{\ell=1}^n (1- m_\ell + (p-1)\sigma_{\ell}^2) \norm{P^{1/2}\Zbf_0 P^{-1/2}}[p, q]^2 \eqsp,
    \end{equation}
    where $\kappa_P = \lambda_{\sf min}^{-1}( P )\lambda_{\sf max}( P )$.
\end{Prop}
\begin{proof}[Proof of \Cref{th:general_expectation} ]
  Let $2 \le q \le p$. Consider the following decomposition $ \Zbf_n = \Y_n \Zbf_{n-1} = (\Y_n - \PE[ \Y_n]) \Zbf_{n-1}+ \PE[\Y_n] \Zbf_{n-1} $,
      Therefore, we obtain for any $n \in\nset$,
      \begin{equation*}
        f_P(\Zbf_n) = \bfA_n + \bfB_n \eqsp, \quad \bfA_n = f_P((\Y_n - \PE[ \Y_n]) \Zbf_{n-1}) \eqsp, \quad \Bbf_n = f_P(\PE[\Y_n]) f_P( \Zbf_{n-1} ) \eqsp,
      \end{equation*}
       where $f_P : \rset^{d \times d } \to \rset^{d \times d}$ is defined for any $B \in \rset^{ d \times d }$ by $f_P(B) = P^{1/2} B P^{-1/2}$.
    Since \(\PE[\bfA_n \vert \bfB_n] = 0\),  \cite[Proposition 4.3]{huang2020matrix} (see \Cref{prop:subquadratic} in  \Cref{sec:techn-supp-results_MP}) implies that 
    \begin{equation}
      \label{eq:bound_decomp_proof_mat_prod}
              \norm{f_P(\Zbf_n)}[p,q]^2 \leq \norm{\bfB_n}[p,q]^2 + (p-1) \norm{\bfA_n}[p,q]^2 \eqsp. 
    \end{equation}
    It remains to bound the two terms on the right hand side. To this end, we use \cite[Theorem 6.20]{hiai:petz:2014} which implies that for any $B_1,B_2 \in \rset^{d \times d}$,
    \begin{equation}
      \label{eq:submultip}
     \norm{B_1 B_2}[p,q]\leq \normop{B_1} \norm{B_2}[p,q]  \eqsp. 
    \end{equation}

    As a result and using that for any $B \in\rset^{d \times d}$,  $\norm{B}[P] = \normop{f_P(B)}$,  and \(\norm{\Y_n - \PE[\Y_n]}[P]  \leq \sigma_{n}\) we get
    \begin{align}
      \nonumber
        \norm{\bfA_n}[p,q] &= \left(\PE\left[ \norm{ f_P(\Y_n - \PE[ \Y_n])  f_P(\Zbf_{n-1})}[p]^q \right]\right)^{1/q} \\
      \label{eq:bound_MP_1_proof}
        &\le  \left(\PE\left[ \normop{\Y_n - \PE [\Y_n]}[P]^q  \norm{f_P(\Zbf_{n-1})}[p]^q  \right]\right)^{1/q} \leq\sigma_{n} \norm{ f_P(\Zbf_{n-1})}[p,q] \eqsp.
    \end{align}
       Similarly, applying $\norm{\PE[\Y_n]}[P]^2  \leq 1 - m_n$
       \begin{align}
               \nonumber
        \norm{\bfB_n}[p,q]^2 &= \left(\PE\left[ \norm{ f_P(\PE[\Y_n])  f_P(\Zbf_{n-1})}[p]^q \right]\right)^{2/q} \\
            \label{eq:bound_MP_2_proof}
        &\le  \left(\PE\left[ \norm{\PE[\Y_n]}[P]^q \norm{f_P(\Zbf_{n-1})}[p]^q \right]\right)^{2/q} \leq (1 - m_n) \norm{f_P(\Zbf_{n-1})}[p,q]^2 \eqsp.
    \end{align}
    
    Combining \eqref{eq:bound_MP_1_proof} and \eqref{eq:bound_MP_2_proof} in \eqref{eq:bound_decomp_proof_mat_prod} yields for any $n \in\nsets$, $\norm{f_P(\Zbf_n)}[p,q]^2 \leq (1 - m_n + (p-1)\sigma_{n}^2)  \norm{f_P(\Zbf_{n-1})}[p,q]^2 \leq \prod_{i=1}^n (1 - m_n + (p-1)\sigma_{n}^2)  \norm{f_P(\Zbf_{0})}[p,q]^2$. 
    The proof is then completed upon using \eqref{eq:submultip}  which implies that $\norm{\Zbf_n}[p,q]= \norm{P^{-1/2}f_P(\Zbf_n) P^{1/2}}[p,q] \leq \sqrt{\kappa_P} \norm{f_P(\Zbf_n)}[p,q]$.
\end{proof}

In order to bound $\ProdBa_{1:n}$ using \Cref{th:general_expectation}, we identify the latter with $\Y_\ell = \Id - \alpha \Am_\ell, \ell \geq 1$, $\Y_0 = \Id$. As $-\bA$ is Hurwitz, applying \Cref{lem:Hurwitzstability} yields $\norm{\PE[\Y_\ell]}[Q]^2 = \norm{\Id - \alpha \bA }[Q]^2 \leq 1 - a \alpha$. Further, \Cref{assum:noise}-\ref{assum:Abounded} ensures that almost surely,
\[
    \norm{\Y_\ell - \PE[\Y_\ell]}[Q] =  \alpha \norm{ \Am_\ell- \bA}[Q] \leq  2 \alpha \sqrt{\qcond} \bConst{A}  = b_{Q} \alpha \eqsp. 
\]
Therefore, \eqref{eq:gen_expectation} holds with $m_\ell = a \alpha$ and  $\sigma_{\ell} = b_Q \alpha$. Noting that as $\norm{\Id}[p] = d^{1/p}$, we obtain the following corollary.

\begin{Corollary}
  \label{cor:norm_Gamma_m_n}
Assume \Cref{assum:noise}-\ref{assum:Abounded}-\ref{assum:Hurwitzmatrices}.
Then, for any $\alpha \in [0, \alpha_{\infty}]$, $2 \le q \le p$, and $n \in \nset$,
\begin{equation}
\label{eq:concentration iid}
\PE^{1/q}\left[ \normop{\ProdBa_{1:n}}^{q} \right]  
\leq  \norm{\ProdBa_{1:n}}[p,q] 
\leq \sqrt{\qcond} d^{1/p} (1 - a \alpha + (p-1) b_Q^2 \alpha^2)^{n/2} \eqsp,
\end{equation}
where $\alpha_\infty$ was defined in \eqref{eq: kappa_def}, and 
\begin{equation}
\label{eq:const_C_p_def}
b_{Q} = 2 \sqrt{\qcond} \bConst{A}\eqsp. 
\end{equation}
\end{Corollary}

Note that \Cref{cor:norm_Gamma_m_n} shows  $\sup_{n\in\nset} \PE[\normop{\ProdBa_{1:n}}^{p}]< \plusinfty$ for any $\alpha \in \ocint{0,\alpha_{p,\infty}}$, where
\begin{equation}
\label{eq:def_alpha_p_infty}
\alpha_{p,\infty} = \alpha_{\infty} \wedge a/(2b_Q^2(p-1))\eqsp.
\end{equation}
This kind of condition relating the choice $\alpha$ with the order $p$ of the moment to be bounded is necessary as illustrated in \Cref{exa:no_p_moments}.
The above corollary further leads to the high-probability bound:
\begin{Corollary}
    \label{corr:concentration_iid}
Assume \Cref{assum:noise}-\ref{assum:Abounded}-\ref{assum:Hurwitzmatrices}.
Then, for any $\alpha \in (0, \alpha_{\infty})$ where $\alpha_\infty$ was defined in \eqref{eq: kappa_def}, $\delta \in (0,1)$ and $n \in \nset$, with probability at least $1 - \delta$,
$$
\normop{\ProdBa_{1:n}} \le   \sqrt{\qcond} \exp \parentheseDeux{-\parentheseLigne{a n \alpha  - \alpha^2 b_Q^2 n}/2 + b_Q \alpha \sqrt{2n\log(d/\delta)}} \eqsp. 
$$
\end{Corollary}
\begin{proof}
The result follows from combining \Cref{cor:norm_Gamma_m_n}  with $p=q$ and \Cref{lem:moments_to_concentr}  in \Cref{sec:techn-supp-results_MP}  applied  with $\sfA = (-\log(\qcond) + a \alpha n + b_Q^2 \alpha^2 n)/2$, $\sfB = \alpha^2 b_Q^2 n/2$ and $\sfC = d$, $p_0=2$, $p_1=+\infty$.
\end{proof}
The result which we obtain in \Cref{corr:concentration_iid} is tight with respect to $\delta$, as illustrated via the following example that continues from \Cref{exa:no_p_moments}.
\begin{Exa*}[Continuation of \Cref{exa:no_p_moments}]
  Consider $\sequence{\theta}[n][\nset]$
  defined by \eqref{eq:lsa} with $\sequence{\Am}[n][\nsets]$ given in
  \eqref{eq:particular_example} and $\bm_n = 0$ for any $n\in\nsets$.
Define
\begin{equation}
  \varphi_q(\alpha) =   q_A \log\parenthese{\frac{1+\alpha}{1-\alpha}} - \log(1+\alpha), \quad \balpha_q = \sup \{\balpha >0 \, : \,  \varphi_q(\alpha) >0,~\forall~ \alpha \in \ooint{0,\balpha}\} \eqsp.
\end{equation}
Note that $  \varphi_q(\alpha) \sim  \alpha(2 q_A-1)$ as $\alpha \downarrow 0$. Therefore since $q_A > 1/2$, $\{\balpha >0 \, : \,  \varphi_q(\alpha) >0 \text{ for any } \alpha \in \ooint{0,\balpha}\} \not = \emptyset$ and  $\balpha_q$ is well-defined.  Consider  also $\tvarphi_q (\alpha) = \varphi_q(\alpha) \log^{-1}[(1+\alpha)/(1-\alpha)]$. Then, we show in \Cref{sec:techn-supp-results_MP} that for any $\bar{\delta} \in \ooint{\rme^{-2n\tvarphi_q(\alpha)},1}$ and $\underline{\delta} \in \oointLigne{\rme^{-n\tvarphi_q^2(\alpha)/(q_A(1-q_A))-2^{-1} \log(n)},1}$,
  \begin{align}
      \label{propo:particular_example_eq_1}
    \proba{\theta_n \geq \exp\parenthese{- \varphi_q(\alpha) n  + \log\parenthese{\frac{1+\alpha}{1-\alpha}}\sqrt{ \frac{ n\log(1/\bar{\delta})} {2} }}} &\leq \bar{\delta} \eqsp, \\
          \label{propo:particular_example_eq_2}
\proba{\theta_n \geq \exp\parenthese{- \varphi_q(\alpha) n  + \log\parenthese{\frac{1+\alpha}{1-\alpha}}\sqrt{n q_A(1-q_A)\log(1/\underline{\delta})+ \frac{n \log(n)}{2}}}} &\geq \underline{\delta}    \eqsp.
  \end{align}
 Note that the bound given by \eqref{propo:particular_example_eq_1} and \eqref{propo:particular_example_eq_2} shows
that the tail distribution associated with $\theta_n$ behaves as a
log-normal one. Indeed, if $\xi$ is a zero-mean one dimensional Gaussian
distribution with unit variance, then an easy computation shows that
for any $\sigma >0$,
$\PP(\rme^{\sigma \xi} \geq t) \sim (2\uppi\sigma^2 )^{-1/2}
\log^{-1}(t) \exp(-(2\sigma^2)^{-1} t^2)$ as $t \to \infty$,
therefore, to have $\PP(\rme^{\sigma \xi} \geq t_{\delta}) \leq \delta$
for a small $\delta>0$, then $t_{\delta}$ has to be of order
$\exp(\sigma \sqrt{\log(1/\delta)})$. 
\end{Exa*}

We conclude the section with a complementary result of \Cref{cor:norm_Gamma_m_n} that does not require \Cref{assum:noise}-\ref{assum:Abounded}:
\begin{Prop}
\label{cor:norm_Gamma_m_n_unbounded}
Assume \Cref{assum:noise}-\ref{assum:Hurwitzmatrices},  $\norm{\Am_1 - \bA} \in \SG(\bConst{A}')$ for some $\bConst{A}' > 0$. Then, for any $\alpha \in (0, \alpha_{\infty})$ where $\alpha_\infty$ was defined in \eqref{eq: kappa_def}, $2 \leq q \leq p$, and $n \in \nset$,
\begin{equation}
\label{eq:concentration iid_unbounded}
\PE^{1/q}\left[ \normop{\ProdBa_{1:n}}^{q} \right]  
\leq  \norm{\ProdBa_{1:n}}[p,q] 
\leq \sqrt{\qcond} d^{1/p} (1 - a \alpha +q (p-1) (b_Q')^2 \alpha^2)^{n/2} \eqsp,
\end{equation}
where $b_Q'  = 2 \sqrt{\qcond}  \bConst{A}'$.
\end{Prop}
The proof is similar to that of \Cref{th:general_expectation} and it can be found in Appendix~\ref{sec:techn-supp-results_MP}.

 \section{Finite-time High-probability Bounds for LSA}
 \label{sec:finite-time-high}

Relying on the results established in \Cref{sec:matrix_concentration} and the decomposition \eqref{eq:LSA_recursion_expanded},
we derive in this section high probability bounds on
$u^{\transpose} \{\theta_n - \thetas\}$ for any $n \in\nset$ and
$u \in \sphere^{d-1}$, where $\sequence{\theta}[n][\nset]$ is defined in \eqref{eq:lsa}. We begin our study with the transient term $\utheta_{n}$ defined in \eqref{eq:LSA_recursion_expanded}. The proof of the following statement is given in \Cref{sec:proof_finite_time_transient}.
\begin{Prop}
\label{prop:transient_term_bound}
Assume \Cref{assum:noise} and let $p_0 \geq 2$. Then, for any $n \in\nsets$, $\alpha \in (0,\alpha_{p_0,\infty})$, where $\alpha_{p_0,\infty}$ is defined in \eqref{eq:def_alpha_p_infty}, $u \in \sphere^{d-1}$ and $\delta \in (0,1)$ it holds with probability at least $1-\delta$ that
\begin{equation*}
|u^\top \ProdBa_{1:n}(\theta_0 - \thetalim)| \leq \sqrt{\qcond}d^{1/p_0} (1-a\alpha/4)^{n} \norm{\theta_0 - \thetas} \delta^{-1/p_0} \eqsp,
\end{equation*}
where $a$ was  defined in \eqref{eq: kappa_def}.
\end{Prop}
\Cref{prop:transient_term_bound} only provides a polynomial high probability bound with respect to $\delta$.
This is due to the fact that only polynomial moments of  $\normop{\ProdBa_{1:n}}$ up to a maximal order are uniformly bounded in the number of iterations $n$. 

We now turn on the fluctuation term  $\vtheta_{n}$ defined in \eqref{eq:LSA_recursion_expanded}. We note that under \Cref{assum:noise}, the sequence $\sequence{\funnoisew}[n][\nset]$ defined in \eqref{eq:LSA-recursion-main} is \iid. From this observation and following  \cite{durmus2021stability}, we consider the decomposition
\begin{equation}
  \label{eq:decompo_vtheta}
  \vtheta_{n} = \alpha \sum_{j=1}^n \ProdBa_{j+1:n} \funnoisew_j = \Jnalpha{n}{0} + \Hnalpha{n}{0} \eqsp,
\end{equation}
where $\sequenceDouble{\Jnalphawn{0} }{\Hnalphawn{0}}[n][\nset]$ are defined by  induction for $n \geq 0$ as:
\begin{equation}
\label{eq:jn0_main}
\begin{array}{ll}
\Jnalpha{n+1}{0}= \left(\Id - \alpha \bA \right) \Jnalpha{n}{0}+ \alpha  \funnoisew_{n+1} \eqsp, & \Jnalpha{0}{0}=0 \eqsp, \\[.1cm]
  \Hnalpha{n+1}{0}= \left( \Id - \alpha \Am_{n} \right) \Hnalpha{n}{0} - \alpha ( \Am_{n+1} - \bA ) J_{n}^{(\alpha,0)} \eqsp, & \Hnalpha{0}{0}=0  \eqsp.
\end{array}
\end{equation}
The latter recurrence can be written as
\begin{equation*}
\Jnalpha{n}{0} = \alpha \sum_{j=1}^{n}  \Ga_{j+1:n} \funnoisew_j \eqsp, \quad \Hnalpha{n}{0} = - \alpha \sum_{j=1}^{n}  \ProdBa_{j+1:n} (\Am_j - \bA) \Jnalpha{j-1}{0} \eqsp.
\end{equation*}
Note that $\Jnalpha{n}{0}$ is a linear statistics of the random variables  $\sequence{\funnoisew}[j][\{1,\ldots,n\}]$ which are centered and \iid~under \Cref{assum:noise}. In our next results, we show that $\Jnalpha{n}{0}$ is the leading term as the stepsize $\alpha \downarrow 0$. Denote for any $n\in\nsets$ and $\alpha >0$, the covariance matrix of $\Jnalpha{n}{0}$ as
\begin{equation}
  \label{eq:def_var_j_n}
 \lineGa_n = \Cov(\Jnalpha{n}{0}) \eqsp. 
\end{equation}
\begin{Prop}
\label{th:HP_bound_J_n_0_rosenthal new}
Assume \Cref{assum:noise}. Then for any $n\in\nsets$,  $\alpha \in (0,\alpha_{\infty}]$, where $\alpha_{\infty}$ is defined in \eqref{eq: kappa_def}, $u \in \sphere^{d-1}$ and $\delta \in (0,1)$, it holds  with probability at least $1 - \delta$,
\begin{equation}
\label{eq:J_n_0_HP_bounds_rosenthal}
\bigl|u^\top \Jnalpha{n}{0}\bigr| < \sfD_{1} \sqrt{  \{u^\top \lineGa_n u \} \log (2/\delta) }+ \alpha \sqrt{1 + \log(1/(a\alpha))} \ConstD_{2} \log^{3/2} (2/\delta) \eqsp,
\end{equation}
where ${\sfD_1} = 60 \sqrt{3} \rme^{4/3}$ and $\sfD_2$ is defined in \eqref{eq:const_D_1_D_2_def}.
\end{Prop}
The proof of \Cref{th:HP_bound_J_n_0_rosenthal new} is postponed to \Cref{subsec:proof_J_n_0_bound}.

We analyze further the covariance associated with $\Jnalpha{n}{0}$  and its dependence with respect to $n$ and $\alpha$. First, note that for any $\alpha \in \ocint{0,\alpha_{2,\infty}}$, $\sequence{\lineGa}[n][\nsets]$ converges to $\alpha \lineGa$  as $n \rightarrow \infty$ where $\lineGa = \alpha \sum_{k=0}^{\infty} G_{1:k} \lineG_{\varepsilon} G_{1:k}^{\transpose}$ is the unique solution of the Ricatti equation 
\begin{equation}
  \label{eq:ricatti}
    \barA \lineGa + \lineGa \barA^{\transpose} - \alpha \barA \lineGa \barA^{\transpose} = \lineG_{\varepsilon} \eqsp, \quad \text{with} \quad \lineG_{\eps} = \PE[\eps_1 \eps_1^{\transpose}] \eqsp.
  \end{equation}
  Indeed, using \Cref{lem:Hurwitzstability}, we easily get that for any $n \geq 0$,
  \begin{equation}
    \label{eq:decomp_lineG_n}
    \norm{\lineGa_n - \alpha \lineGa} \leq \alpha^2  \sum_{k >n} \norm{G_{1:k}}^2 \norm{\lineG_{\eps}} \leq \alpha a^{-1}\qcond   \norm{\lineG_{\eps}} (1-\alpha a)^n \eqsp. 
  \end{equation}
  We now give an expansion of  $\lineGa$ with respect to $\alpha$. It is well-known that as $\alpha \downarrow 0$, $\lineGa$ converges to $\lineG$, the unique solution of the Lyapunov equation (see \cite[Lemma~9.1]{poznyak:control})
\begin{equation}
  \label{eq:def_lineG}
  \bA \lineG + \lineG \bA^{\transpose} = \lineG_{\varepsilon}\eqsp. 
\end{equation}
Our next result states this convergence is of the order of the stepsize $\alpha$. 
\begin{Prop}
\label{prop:Sigma_alpha_expansion}
Assume that \Cref{assum:noise}-\ref{assum:Hurwitzmatrices} holds. 
Then, for any $\alpha \in (0,\alpha_{\infty}]$, where $\alpha_{\infty}$ is defined in \eqref{eq: kappa_def},
\[
  \norm{\lineGa -   \lineG}[Q] \leq \alpha a^{-1}\|\bA \lineG \bA^\top\|_{Q}   \eqsp,
\]
where $\lineGa$ and $\lineG$ are defined in  \eqref{eq:ricatti} and \eqref{eq:def_lineG} respectively and  $a$ is given in
 \eqref{eq: kappa_def}.
\end{Prop}
The proof is given in \Cref{subsec:proof_var_J_n_0}.
The last step in bounding $\vtheta_n$ is to consider $\Hnalpha{n}{0}$. We proceed similarly to \eqref{eq:jn0_main} and consider the decomposition $\Hnalpha{n}{0} = \Jnalpha{n}{1} + \Hnalpha{n}{1}$,
where $\sequenceDouble{\Jnalphawn{1} }{\Hnalphawn{1} }[n][\nset]$ are defined by  induction for $n \geq 0$ as:
\begin{equation}
\begin{array}{ll}
\label{eq:expansion_recur_gen}
\Jnalpha{n+1}{1} = (\Id - \alpha \bA) \Jnalpha{n}{1} - \alpha (\Am_{n+1} - \bA) \Jnalpha{n}{0}, & \Jnalpha{0}{1}=0 \eqsp, \\[.1cm]
\Hnalpha{n+1}{1} = (\Id - \alpha \Am_{n+1} ) \Hnalpha{n}{1} - \alpha (\Am_{n+1} - \bA) \Jnalpha{n}{1}, & \Hnalpha{0}{1} = 0\eqsp.
\end{array}
\end{equation}
In our next results, we bound each term of this decomposition separately. 
\begin{Prop}
\label{prop:J_n_1_bound_new}
Assume \Cref{assum:noise} and  let $p_0 \geq 2$. Then, for any $n \in\nset$, $\alpha \in (0,\alpha_{p_0,\infty})$, where $\alpha_{p_0,\infty}$ is defined in \eqref{eq:def_alpha_p_infty},  $u \in \sphere^{d-1}$ and 
$\delta \in (0,1/2)$, with probability at least $1 - 2\delta$,  it holds 
\begin{equation}
\label{eq:HPB-J1}    
\bigl|u^\top \Jnalpha{n}{1}\bigr| < \rme \sfD_3 \alpha \log^2(1/\delta)\eqsp, \quad \bigl| u^\top \Hnalpha{n}{1} \bigr| < \ConstD_4 \alpha p_0^2 \delta^{-1/p_0}\eqsp,
\end{equation}
where $\sfD_{3}$ and $\ConstD_4$ are given in \eqref{eq: Jn1 bound} and \eqref{eq:moment_bound_H_n_1}, respectively.
\end{Prop}
The proof of \Cref{prop:J_n_1_bound_new} is postponed to \Cref{subsec:proof_J_n_1_bound}. Now we are ready to combine the previous bounds and to state the main result of this section.
\begin{theorem}
\label{th:HP_error_bound_iid}
Assume \Cref{assum:noise} and  let $p_0 \geq 2$. Then, for any $n \in\nset$, $\alpha \in (0,\alpha_{p_0,\infty})$, where $\alpha_{p_0,\infty}$ is defined in \eqref{eq:def_alpha_p_infty},  $u \in \sphere^{d-1}$ and 
$\delta \in (0,1/4)$, with probability at least $1 - 4\delta$,  it holds 
{\small
  \begin{equation}
    \label{eq:th:HP_error_bound_iid}
\alpha^{-1/2} |u^\top (\theta_{n}-\thetalim)| < \sfD_{1} \sqrt{ \{u^{\top} \lineGa u\} \log(2/\delta) } + \alpha^{1/2} q^{(1)}(\alpha,\delta) + (1-a\alpha/4)^{n} \Delta^{(1)}(\alpha,\delta)\eqsp,
\end{equation}
}
where  $\lineGa$ is the unique solution of \eqref{eq:ricatti}, $\sfD_{1} = 60 \sqrt{3} \rme^{4/3}$, $a$ is defined in  \eqref{eq: kappa_def},
{\small
\begin{equation}
\label{eq:remainder_definition_A1}
\begin{split}
q^{(1)}(\alpha,\delta) &=  \bigl(\rme \sfD_3 \log^2(1/\delta) +  \sqrt{ 1 + \log(1/a\alpha) } \ConstD_{2} \log^{3/2}(2/\delta)\bigr) + \ConstD_4 p_0^2 \delta^{-1/p_0} \eqsp, \\
\Delta^{(1)}(\alpha,\delta) &= \sfD_{1}\sqrt{a^{-1}\qcond\normop{\lineG_{\varepsilon}} \log(2/\delta) } + \sqrt{\qcond} d^{1/p_0}\normop{\theta_0 - \thetalim}\alpha^{-1/2} \delta^{-1/p_0} \eqsp,
\end{split}
\end{equation}
}
where $\qcond$  and $\lineG_{\varepsilon}$ are  defined in \eqref{eq: kappa_def} and \eqref{eq:ricatti}
respectively.
\end{theorem}
\begin{proof}
The proof follows from  the  decomposition 
\begin{equation*}
u^\top (\theta_{n} - \thetalim) = u^\top \ProdBa_{1:n} (\theta_{0} - \thetalim) + u^\top \Jnalpha{n}{0} + u^\top \Jnalpha{n}{1} + u^\top \Hnalpha{n}{1}\eqsp,
\end{equation*}
where $\Jnalpha{n}{0}$, $\Jnalpha{n}{1}$ and $\Hnalpha{n}{1}$ are
defined in \eqref{eq:jn0_main}-\eqref{eq:expansion_recur_gen}, the
union bound and \Cref{prop:transient_term_bound},
\Cref{th:HP_bound_J_n_0_rosenthal new}, \eqref{eq:decomp_lineG_n} and \Cref{prop:J_n_1_bound_new}.
\end{proof}
We now discuss the high probability bound \eqref{eq:th:HP_error_bound_iid}. First, note that the term $\Delta^{(1)}(\alpha,\delta)$, and in particular the initial condition vanishes exponentially fast in the number of iterations $n$. In addition,   $q^{(1)}(\alpha,\delta)$ and $\Delta^{(1)}(\alpha,\delta)$ are of order $\delta^{-1/p_0}$ as $\delta \to 0$ and therefore \eqref{eq:th:HP_error_bound_iid} provides polynomial high probability bounds on LSA. However, this conclusion is expected as illustrated in \Cref{exa:no_p_moments}.
Finally, the discussion of \eqref{eq:th:HP_error_bound_iid} with respect to $\alpha$ is postponed to  the next section.

Under \Cref{ass:contraction} we can provide a better bound for $\Hnalpha{n}{1}$. 
\begin{Prop}
\label{prop:H_n_1_bound_contractive}
Assume \Cref{assum:noise} and \Cref{ass:contraction}. Then, for any $n \in\nset$, $\alpha \in (0,\alpha_{\infty} \wedge \talpha_{\infty})$, where $\alpha_{\infty}$ is defined in \eqref{eq: kappa_def},  $u \in \sphere^{d-1}$ and 
$\delta \in (0,1/2)$, with probability at least $1 - 2\delta$,  it holds
\begin{equation}
\label{eq:HPB-H1-contractive}    
\bigl|u^\top \Jnalpha{n}{1}\bigr| < \rme \sfD_3 \alpha \log^2(1/\delta)\eqsp, \quad \bigl|u^\top \Hnalpha{n}{1}\bigr| < \rme \ConstD_5 \alpha \log^2(1/\delta)\eqsp,
\end{equation}
where $\sfD_{3}$ and  $\ConstD_5$ are  given in \eqref{eq: Jn1 bound} and \eqref{eq:const_D_5_definition} respectively.
\end{Prop}
As a result, we can establish exponential high probability bounds with respect to $\delta$.
\begin{theorem}
\label{th:HP_error_bound_contractive}
Assume \Cref{assum:noise} and \Cref{ass:contraction}. Then, for any $n \in\nset$, $\alpha \in (0,\alpha_{\infty} \wedge \talpha_{\infty})$, $u \in \sphere^{d-1}$ and 
$\delta \in (0,1/4)$, with probability at least $1 - 4\delta$,  it holds 
{\small
\begin{equation*}
\alpha^{-1/2} |u^\top (\theta_{n} - \thetalim)| < \sfD_{1} \sqrt{ \{u^{\top} \lineGa u\} \log(2/\delta) } + \alpha^{1/2}q^{(2)}(\alpha,\delta) + (1-\alpha \ta)^{n/2} \Delta^{(2)}(\alpha,\delta)\eqsp,
\end{equation*}
}
where $\sfD_{1} = 60 \sqrt{3} \rme^{4/3}$, $\lineGa$ is solution of \eqref{eq:ricatti}, 
{\small
\begin{equation}
\label{eq:remainder_definition_A2}
\begin{split}
q^{(2)}(\alpha,\delta) &= \rme(\sfD_3 + \ConstD_{5})\log^2(1/\delta) + \sqrt{ 1 + \log(1/\ta\alpha)} \ConstD_{2} \log^{3/2}(2/\delta) \eqsp, \\
\Delta^{(2)}(\alpha,\delta) &= \sfD_{1}\sqrt{\ta^{-1}\tqcond\normop{\lineG_{\varepsilon}} \log(2/\delta)} + \tqcond^{1/2}\norm{\theta_0 - \thetas}\alpha^{-1/2}\,\eqsp,
\end{split}
\end{equation}
}
where $\lineG_{\varepsilon}$ is defined in \eqref{eq:ricatti}.
\end{theorem}
\begin{proof}
The proof follows the lines of \Cref{th:HP_error_bound_iid} with \Cref{prop:H_n_1_bound_contractive} used instead of \Cref{prop:J_n_1_bound_new}.
\end{proof}

\section{Optimality of the derived bounds with respect to \texorpdfstring{$\alpha$: analysis of $(\theta_n)_{n\in\nset}$ as a Markov chain}{alpha}}
\label{sec:optim-deriv-bounds}

In this section, we study the sequence $\sequence{\theta}[n][\nset]$ defined in \eqref{eq:lsa} as a Markov chain. This perspective will allow us to show that the bounds that we derived in
\Cref{th:HP_error_bound_iid}
are near-Berstein high probability bounds with respect to the stepsize $\alpha$.  
Denote by $R_{\alpha}$ the Markov kernel associated with
$\sequencen{\theta}$.  First, we show that if $\alpha$ is small enough
then $R_{\alpha}$ is geometrically ergodic with respect to the
Wasserstein distance of order $2$ denoted by $W_2$ and give a
representation of its stationary distribution as an infinite sum.

\begin{theorem}
  \label{theo:convergence_mc_wasserstein}
  Assume \Cref{assum:noise}. Then,   for any $\alpha\in \ooint{0,\alpha_{2,\infty}}$, where $\alpha_{2,\infty}$ is defined in \eqref{eq:def_alpha_p_infty},  $R_{\alpha}$ admits a unique stationary distribution $\pi_{\alpha} \in \Pens_2(\rset^d)$ and for any $n \in\nset$,
  \begin{equation}
    \label{eq:conv_wasser_pi_alpha}
    W_2^2 (\updelta_{\theta} R_{\alpha}^n, \pi_{\alpha}) \leq \sqrt{\qcond d  (1 - a \alpha/2)^{n} } \int_{\rset^d} \norm{\ttheta-\theta}^2 \rmd \pi_{\alpha} (\ttheta)\eqsp.
  \end{equation}
  Further, if $\sequenceDouble{\Am}{\bm}[k][\nsetm]$ is any sequence of \iid~random variables with the same distribution as $(\Am_1,\bm_1)$, then the following limit exists almost surely and in $\mrl^2$ and has  distribution
  $\pi_{\alpha}$:
  \begin{equation}
    \label{eq:def_theta_infty_alpha}
    \thetalima = \lim_{n \to \minfty} \theta_{n}^{(\alpha,\leftarrow)}  \eqsp, \quad \theta_{n}^{(\alpha,\leftarrow)} = \alpha \sum_{k=n}^1 \Gamma_{k:0} \bm_{k-1} \eqsp, \quad \Gamma_{k:0} = \prod_{i=k}^0 (\Id_d - \alpha \Am_i) \eqsp.
  \end{equation}
\end{theorem}
The proof is postponed to \Cref{sec:proof-crefth_mc_wasser}.  Based on
\Cref{th:HP_error_bound_iid}, we easily get concentration bounds for
the family of distributions $\{\pi_{\alpha}\, :\, \alpha \in \ooint{0,\alpha_{2,\infty}} \}$ around $\thetas$.

\begin{theorem}
\label{th:HP_error_bound_iid_statio}
Assume \Cref{assum:noise} and  let $p_0 \geq 2$. Then, for any $\alpha \in (0,\alpha_{p_0,\infty})$,  where $\alpha_{p_0,\infty}$ is defined in \eqref{eq:def_alpha_p_infty},
$u \in \sphere^{d-1}$ and $\delta \in (0,1/4)$, with probability at
least $1 - 4\delta$, it holds
{\small
  \begin{equation}
    \label{eq:th:HP_error_bound_iid_alpha}
\alpha^{-1/2} |u^\top (\thetalima-\thetalim)| <  \sfD_{1}\sqrt{\{u^{\top} \lineG u\} \log(2/\delta) } + \alpha^{1/2} [  a^{-1/2}\|\bA \lineG \bA^\top\|_{Q}^{1/2} + q^{(1)}(\alpha,\delta) ]\eqsp,
\end{equation}
}
where $\lineG$ is the unique solution  of \eqref{eq:def_lineG}, $\sfD_{1} = 60\sqrt{3}\rme^{2/3}$, $a$ is defined in  \eqref{eq: kappa_def}, and $q^{(1)}(\alpha, \delta)$ in \eqref{eq:remainder_definition_A1}.
\end{theorem}

\begin{proof}
The proof follows from \Cref{th:HP_error_bound_iid}, the Portmanteau theorem \cite[Theorem 13.16]{klenke:2013}, and the fact that convergence in $W_2$ implies weak
convergence.
\end{proof}
Our results is only polynomial in $\delta$ and we cannot expect improving this dependency as illustrated in \Cref{exa:no_p_moments} for fixed $\alpha$.
The leading term in \eqref{eq:th:HP_error_bound_iid_alpha}  as $\alpha \downarrow 0$ is $\sqrt{ \sfD_{1} \{u^{\top} \lineG u\} }$.  
In our next result, we establish a central limit theorem for the
family $(\thetalima)_{\alpha
  \in\ocint{0,\alpha_{2,\infty}}}$ where $\lineG$ plays the role of the asymptotic covariance matrix. As a result, \eqref{eq:th:HP_error_bound_iid_alpha} is a Bernstein-type high probability bound with respect to $\alpha$ and therefore \eqref{eq:th:HP_error_bound_iid_alpha} is sharp.
Define for any $\alpha \in \ocint{0,\alpha_{2,\infty}}$,
\begin{equation}
  \label{eq:tthetalima}
\tthetalima =   \alpha^{-1/2}\{\thetalima - \thetas\}\eqsp.
\end{equation}
\begin{theorem}
  \label{theo:tcl_theta_inf_alpha}
  Assume \Cref{assum:noise}. Then, the family $\{\tthetalima \, : \, \alpha \in \ocint{0,\alpha_{2,\infty}}\}$ converges in law as $\alpha \downarrow 0$ to a zero-mean Gaussian random variable with covariance matrix $\lineG$ defined by \eqref{eq:def_lineG}. 
\end{theorem}
Note that this result was established in \cite[Theorem 1]{pflug:1986} for general stochastic approximation schemes but under stronger conditions on the sequence $\sequence{\varepsilon}[n][\nsets]$. In particular, it is assumed that the distribution of $\varepsilon_1$ admits a density with respect to the Lebesgue measure. We relax this condition and provide a new proof for this result. In particular, our strategy to establish \Cref{theo:tcl_theta_inf_alpha} is to
consider the decomposition \eqref{eq:decompo_vtheta} of 
$\sequence{\theta}[n][\nset]$ with $\theta_0 =0 $,  since in such case $\theta_n = \vtheta_n$ for any $n\in\nset$.
Define $\sequence{J^{(\alpha, \leftarrow)}}[n][\nsetm]$ by 
\begin{equation}
  \label{eq:def_statio_truncated}
J^{(\alpha,\leftarrow)}_n =  \alpha \sum_{k=n}^{1}  G_{k:0} \funnoisew_{k-1} \eqsp, \qquad   G_{k:0} = \prod_{i=k}^0(\Id - \alpha \bA) \eqsp. 
\end{equation}
Note that for any $n \in \nset$, $\theta_{-n+1}^{(\alpha,\leftarrow)}$ has the same distribution as $\theta_{n}^{(\alpha)}$ starting from
$\theta_0=0$ and $\Jnalpha{n}{0}$ as
$J^{(\alpha,\leftarrow)}_{-n+1}$. In contrast to $\Jnalpha{n}{0}$,
$J^{(\alpha,\leftarrow)}_{-n+1}$ admits a limit in $\mrl^2$ and almost
surely denoted by $J^{(\alpha,\leftarrow)}_{\infty}$.  Then, we get
for any $u \in \sphere^{d-1}$, $\alpha \in (0, \alpha_{2,\infty}]$, bounded and Lipschitz functions $f: \rset \to \rset$, with Lipschitz constant smaller than $1$, by the Lebesgue dominated convergence theorem
\begin{align}
  \nonumber
  &  \abs{  \PE[f(u^{\transpose}\ttheta_{\infty}^{(\alpha,\leftarrow)})]  - \PE[f(\alpha^{-1/2} u^{\transpose} J_{\infty}^{(\alpha,\leftarrow)})]} \\
  \nonumber
  &=
    \lim_{n \to \plusinfty}  \abs{  \PE[f( \alpha^{-1/2} u^{\transpose}[\theta_{-n+1}^{(\alpha,\leftarrow)}-\thetas])] - \PE[f(\alpha^{-1/2}u^{\transpose} J_{-n+1}^{(\alpha,\leftarrow)})]}\\
    \nonumber
  \label{eq:bound_h_n_prood_clt}
  & = \lim_{n \to \plusinfty}  \abs{  \PE[f( \alpha^{-1/2} u^{\transpose}[\theta_{n}^{(\alpha)}-\thetas])] - \PE[f(\alpha^{-1/2} u^{\transpose}\Jnalpha{n}0)]} \leq  \limsup_{n\to \plusinfty} \PE[\abs{\alpha^{-1/2} u^\transpose \Hnalpha{n}{0} }] \eqsp. 
\end{align}
Using  the decomposition $\Hnalpha{n}{0} = \Jnalpha{n}{1} + \Hnalpha{n}{1}$,
where $\sequenceDouble{\Jnalphawn{1} }{\Hnalphawn{1} }[n][\nset]$ are defined in \eqref{eq:expansion_recur_gen}  and plugging the bounds provided by \Cref{prop:J_n_1_bound_new_app} and \Cref{prop:H_n_1_bound_new_app} in Appendix~\ref{subsec:proof_J_n_1_bound} shows that
\begin{equation*}
  \limsup_{\alpha \to 0} \abs{  \PE[f(u^{\transpose}\ttheta_{\infty}^{(\alpha,\leftarrow)})]  - \PE[f(\alpha^{-1/2} u^{\transpose} J_{\infty}^{(\alpha,\leftarrow)})]} = 0 \eqsp.
\end{equation*}
Therefore by the Cramer Wold device and the Portmanteau theorem \cite[Theorem 13.16]{klenke:2013}, \Cref{theo:tcl_theta_inf_alpha} follows from the next result.
\begin{Prop}
  \label{propo:clt_J_n_0}
  Assume \Cref{assum:noise}. Then, for any $u \in \sphere^{d-1}$,
  $\{\alpha^{-1/2} u^{\transpose} J_{\infty}^{(\alpha,\leftarrow)} \,
  :\, \alpha \in \ocint{0,\alpha_{2,\infty}}\}$ converges in
  distribution to the zero-mean Gaussian distribution with variance
  $u^{\transpose} \lineG u$ where $\lineG$ is given in
  \eqref{eq:def_lineG}.
\end{Prop}
The proof is postponed to \Cref{sec:proof-crefpr_clt_j}.


\section{Conclusion}
\label{sec: conclusion}
In this paper, we provided a novel non-asymptotic analysis of LSA algorithms with fixed stepsize. For any $\delta \in (0,1)$, we obtain bounds on the sequence $\sequenceD{\norm{\theta_n - \thetas}}[n \in \nset]$ that holds with probability at least $1-\delta$. The bounds are proven to be tight with respect to the stepsize, and we show that these bounds necessarily have polynomial dependency in $\delta$.
Importantly, our results do not require the matrices ${\bf A}_n$ to be symmetric but only Hurwitz, which enables one to apply them to various scenarios such as reinforcement learning.
Future work includes extending our high probability bounds to a larger panel of random noise, e.g., with heavy tailed distribution, Markovian dependency, as well as Polyak-Ruppert averaging.

\bibliographystyle{abbrvnat}
\bibliography{references}

\appendix

\section{Proofs of \texorpdfstring{\Cref{sec:lsa}}{Section 2}}

\subsection{Proofs of \texorpdfstring{\Cref{lem:Hurwitzstability}}{Proposition 1}}
\label{sec:proofs-crefl_hurwitz_proof}
The existence and uniqueness of $Q$ follows from \cite[Lemma 9.1, p. 140]{poznyak:control}.
Regarding the second statement, note that for any  $x \in \rset^d\setminus\{0\}$, we have
\[
\frac{x^\top (\Id - \alpha \bA)^\top Q (\Id - \alpha \bA) x}{x^\top Q x}
=1 - \alpha \frac{\norm{x}^2}{x^\top Q x} + \alpha^2 \frac{x^\top \bA^\top Q \bA x}{x^\top Q x} \eqsp. 
\]
Hence, we get that for all $\alpha \in [0, \alpha_{\infty}]$,
\begin{align}
1 - \alpha \frac{\norm{x}^2}{x^\top Q x} + \alpha^2
\frac{x^\top \bA^\top Q \bA x}{x^\top Qx}
&\leq 1 - \alpha  \normop{Q}^{-1} + \alpha^2 \normop{\bA}[Q]^2 \leq 1 - (1/2) \normop{Q}^{-1} \alpha \,. \nonumber
\end{align}
The proof is completed using that for any matrix $\bA \in \rset^{d \times d}$, $\norm{\bA}[Q] \leq \qcond^{1/2} \norm{\bA}$.

\subsection{Proof for \texorpdfstring{\Cref{exa:no_p_moments}}{Example 1}}
\label{sec:proof-example_station_no_p_moments}
  The existence and uniqueness of the stationary distribution
  $\pi_{\alpha}$ is a consequence of \Cref{theo:convergence_mc_wasserstein} noting that \Cref{assum:noise} is satisfied for the particular case that we consider. We now show the second statement. Let $\alpha \in\ooint{0,\alpha_{2,\infty}}$.  First, note that since $\bm_1$ is a  zero-mean Gaussian random variables with unit
  variance independent of $\Am_1$, we have for any $p \geq 1$,
  \begin{multline*}
    \PE[\theta_1^{2p}] = \sum_{k=0}^{2p} \binom{2p}{k} \PE[ \theta_0^{2p-k} ] \PE[(1-\alpha \Am_1)^{2p-k}] \PE[\bm_1^k] \\= \sum_{k=0}^{p} \binom{2p}{2k} \PE[ \theta_0^{2(p-k)} ] \PE[(1-\alpha \Am_1)^{2(p-k)}] \PE[\bm_1^{2k}] \geq \PE[ \theta^{2p}_0 ] \PE[(1-\alpha \Am_1)^{2p}] \eqsp. 
  \end{multline*}
  This shows that taking $\theta_0$ with distribution $\pi_{\alpha}$ that if $\int_{\rset} \abs{\theta}^{2p} \rmd \pi_{\alpha}(\theta) < 
  \plusinfty$, then it is necessary that $\PE[(1-\alpha \Am_1)^{2p}] \leq 1$. However, 
  using that
  $\PE[(1-\alpha \Am_1)^{2p}] = \{q_{A}(1-\alpha)^{2p}+(1-q_{A})(1+\alpha)^{2p} \}$ and $(1-\alpha)^{2p} \geq 1-2\alpha p$ and $(1+\alpha)^{2p} \geq 1+2\alpha p  + 2p(2p-1)\alpha^2/2$, we get for any $p \geq 1$,
  $ \PE[(1-\alpha \Am_1)^{2p}] \geq \{1-2p\alpha[(2q_{A}-1) -(2p-1)\alpha(1-q_{A})/2]\}$, therefore $\PE[(1-\alpha \Am_1)^{2p}] \leq 1$ does not hold for $2 p > \bar{p}_{q,\alpha} = 1+2(2q_{A}-1)/[\alpha(1-q_{A})]$.

\section{Technical and supporting results for \texorpdfstring{\Cref{sec:matrix_concentration}}{Section 3}}
\label{sec:techn-supp-results_MP}
\begin{Prop}[\protect{\cite[Proposition 4.3]{huang2020matrix}}]
\label{prop:subquadratic}
    Consider two random matrices \(\X,\Y \in \rset^{d \times d}\) that satisfy \(\PE[\Y | \X] = 0\). Then for \(2 \leq q \leq p\),
    \[
        \norm{\X + \Y}[p,q]^2 \leq  \norm{\X}[p,q]^2 + C_p \norm{\Y}[p,q]^2 \eqsp,
    \]
    where \(C_p = p-1\).
  \end{Prop}

  \begin{lemma}\label{lem:moments_to_concentr}
Let $\sfA\in\rset$, $\sfB>0$, $\sfC\geq 1$, $p_0,p_1\in \rset$ such that $1\leq p_0 \leq p_1 < \plusinfty$, and $X$ a real random variable  satisfying, for any $p\in[p_0,p_1]$,
\begin{equation}
  \label{eq:1:lem:moments_to_concentr}
\PE[|X|^p] \leq \sfC \exp(-\sfA p + \sfB p^2)\,.
\end{equation}
Then, for all $\delta\in(0,1]$, we have, with probability at least $1-\delta$,
\begin{equation}
|X| \leq \exp\left(-\sfA + \sfB p_0 + 2\sqrt{\sfB\log(\sfC/\delta)} + \log(\sfC/\delta)/p_1\right) \eqsp,
\end{equation}
with the convention $c/\infty = 0$ for $c >0$. In addition if \eqref{eq:1:lem:moments_to_concentr} is satisfied for any $p\geq p_0$, then  with probability at least $1-\delta$,
\begin{equation*}
    \label{eq:2:lem:moments_to_concentr}
|X| \leq \exp\left(-\sfA + \sfB p_0 + 2\sqrt{\sfB\log(\sfC/\delta)} \right) \eqsp.
\end{equation*}
\end{lemma}
\begin{proof}
Note that by the monotone convergence theorem,  it is sufficient to show \eqref{eq:1:lem:moments_to_concentr}.  By Markov's inequality, we have, for any $t>0$ and $p\in[p_0,p_1]$,
\begin{equation}
\P(|X| \geq t) \leq \PE[|X|^p] / t^p \leq \sfC\exp\left(-p(\log(t) + \sfA - \sfB p)\right)\,.
\end{equation}
Taking $t = \exp(-\sfA + 2 \sfB a^*)$ for $a^{*} \in \rset$ and maximizing over $p \in\ccint{p_0,p_1}$, we obtain 
\begin{equation}
\P(|X| \geq \exp(-\sfA + 2 \sfB a^*)) \leq \sfC \exp\left(- \sfB p(2a^* - p)\right) \leq \sfC \exp\left(- \sfB \phi(a^*)\right) \eqsp,
\end{equation}
where
\begin{equation*}
  \phi(a^*) = \max_{p\in[p_0,p_1]} p(2a^* - p) = (2p_0a^*-p_0^2)\1_{\ocintLigne{-\infty,p_0}}(a^*)  +  (a^*)^2  \1_{\oointLigne{p_0,p_1}}(a^*)+  (2p_1 a^*-p_1^2)\1_{\cointLigne{p_1,\plusinfty}}(a^*) \eqsp.
\end{equation*}
Note that for any $t \in \rset$, the inverse of $\phi$ is given by
\begin{equation*}
  \phi^{\leftarrow}(t) =  \frac{p_0^2+t}{2p_0}\1_{\ocintLigne{-\infty,p_0^2 }}(t)  +   t^{1/2} \1_{\oointLigne{p_0^2,p_1^2}}(t) +  \frac{p_1^2+t}{2p_1} \1_{\cointLigne{ p_1^2,\plusinfty}}(t)  \eqsp.
\end{equation*}
For $\delta >0$, taking $a^{*}_{\delta} = \phi^{\leftarrow}(\log(\sfC/\delta) / \sfB)$ gives
\begin{equation*}
  \P(|X| \geq \exp[-\sfA + 2 \sfB   \phi^{\leftarrow}(\log(\sfC/\delta) / \sfB)]) \leq \delta \eqsp. 
\end{equation*}
The proof then follows from the fact that for any $t \in\rset$, $\phi^{\leftarrow}(t) \leq p_0/2 + \sqrt{t} + t/(2p_1)$.
\end{proof}

\subsection*{Proof of \eqref{propo:particular_example_eq_1} and \eqref{propo:particular_example_eq_2}}

  Let $\alpha \in \ooint{0,\balpha_q}$. 
Note that by definition of $\sequence{\theta}[n][\nset]$ with \eqref{eq:particular_example}, for any $n \in\nset$,
\begin{equation*}
\theta_n  = (1-\alpha)^{N_n}(1+\alpha)^{n-N_n} \eqsp, \quad \text{ where $N_n = \sum_{k=1}^n \1_{\{1\}}(Z_k)$ } \eqsp.
\end{equation*}
Then, for any $\beta >0$, we get
\begin{align}
  \nonumber
  \proba{\theta_n \geq \rme^{-\alpha \beta n }} &= \proba{\log (\theta_n) \geq - \alpha \beta n } = \proba{N_n \log\parenthese{\frac{1-\alpha}{1+\alpha}} \geq -\alpha\beta n - n \log(1+\alpha)}
  \\
    \nonumber
                                           & = \proba{N_n \leq n \log^{-1}\parenthese{\frac{1+\alpha}{1-\alpha}}\defEns{\alpha \beta + \log(1+\alpha)}} \\
  \label{eq:particular_example_3}
    &  = \proba{N_n -q_An  \leq -n \parentheseDeux{q_A- \log^{-1}\parenthese{\frac{1+\alpha}{1-\alpha}}\defEns{\alpha \beta + \log(1+\alpha)}}} \eqsp.
\end{align}

Let $  \beta_{\alpha,q} = \alpha^{-1} \parentheseDeux{q_A \log\defEns{(1+\alpha)/(1-\alpha)} - \log(1+\alpha)}$.
Note that with the condition, $\alpha\in\ooint{0,\balpha_q}$, $\beta_{\alpha,q} >0$ and therefore for any $\beta \in\ooint{0,\beta_{\alpha,q}}$,
\begin{equation}
  \label{eq:def_w_alpha_beta}
  x_{\alpha,\beta}  = \parentheseDeux{q_A- \log^{-1}\parenthese{\frac{1+\alpha}{1-\alpha}}\defEns{\alpha \beta + \log(1+\alpha)}} \in \ooint{0,\tvarphi_q(\alpha)} \eqsp,
\end{equation}
We now show \eqref{propo:particular_example_eq_1}. From \eqref{eq:particular_example_3}, it follows using Hoeffding's inequality that for any $\beta \in\ooint{0,\beta_{\alpha,q}}$,
\begin{equation}
\label{eq:ineq_proba_theta_n}
  \proba{\theta_n \geq \rme^{-\alpha \beta n }}    \leq \rme^{-2n x_{\alpha,\beta}^2}\eqsp.
\end{equation}
Hence, for $\bar{\delta} \in \oointLigne{\rme^{-2n\tvarphi_q^2(\alpha)},1}$, there exists $x \in \ooint{0,\tvarphi_q(\alpha)}$ such that $\rme^{-2n x^2} = \bar{\delta}$ given by $x = \sqrt{\log(1/\bar{\delta})/2n}$, which corresponds by \eqref{eq:def_w_alpha_beta} to
\begin{equation*}
 \beta = \alpha^{-1}\defEns{q_A-\sqrt{\log(1/\bar{\delta})/ (2n) }}\log\parenthese{\frac{1+\alpha}{1-\alpha}} - \alpha^{-1}\log(1+\alpha) \in\ooint{0,\beta_{\alpha,q}} \eqsp.
\end{equation*}
This completes the proof of    \eqref{propo:particular_example_eq_1}  using \eqref{eq:ineq_proba_theta_n}.

We now show \eqref{propo:particular_example_eq_2}. Using \cite[Lemma 4.7.2]{ash:1966} and \eqref{eq:particular_example_3}, it holds that for any $\beta \in\ooint{0,\beta_{\alpha,q}}$,
\begin{equation}
  \label{eq:particular_example_4}
 \proba{\theta_n \geq \rme^{-\alpha \beta n }} \geq \exp(-n \KLs(q_A-x_{\alpha,\beta} | q_A) - 2^{-1} \log(n)) \eqsp,
\end{equation}
where for any $\tilde{q} \in \ooint{0,1}$,
\begin{equation*}
  \KLs(\tilde{q} | q_A) = \tilde{q} \log(\tilde{q}/q_A) + (1-\tilde{q})\log((1-\tilde{q})/(1-q_A)) \eqsp. 
\end{equation*}
Note that for any $\tilde{q} \in \ooint{0,1}$, $\tilde{q} \leq q_A$, using $\log(1+t) \leq t$ for any $t >-1$, we get
\begin{equation*}
  \KLs(\tilde{q} | q_A) \leq (q_A-\tilde{q})^2/(q_A(1-q_A))\eqsp.
\end{equation*}
Therefore, plugging this result into \eqref{eq:particular_example_4} yields for any $\beta \in\ooint{0,\beta_{\alpha,q}}$,
\begin{equation}
  \label{eq:particular_example_5}
 \proba{\theta_n \geq \rme^{-\alpha \beta n }} \geq \exp(-n x_{\alpha,\beta}^2/ (q_A(1-q_A)) - 2^{-1} \log(n)) \eqsp.
\end{equation}
Hence, for $\underline{\delta} \in \oointLigne{\rme^{-n\tvarphi_q^2(\alpha)/(q_A(1-q_A))-2^{-1} \log(n)},1}$
there exists $x \in \oointLigne{0,\tvarphi_q(\alpha)}$ such that $\rme^{-n x^2 / (q_A(1-q_A)) - 2^{-1} \log(n)} = \underline{\delta}$, given by $x = \sqrt{2^{-1} \log(n)+q_A(1-q_A)\log(1/\underline{\delta})/n}$, which corresponds by \eqref{eq:def_w_alpha_beta} to
\begin{equation*}
  \beta =  \alpha^{-1}\{q_A-\sqrt{2^{-1} \log(n)+q_A(1-q_A)\log(1/\underline{\delta})/n}\}\log\parenthese{\frac{1+\alpha}{1-\alpha}} - \alpha^{-1}\log(1+\alpha) \eqsp.
\end{equation*}
This completes the proof of    \eqref{propo:particular_example_eq_2}  using \eqref{eq:particular_example_5}.

\begin{proof}[Proof of \Cref{cor:norm_Gamma_m_n_unbounded}] It suffices to repeat the argument of \Cref{cor:norm_Gamma_m_n}. We need a version of \Cref{th:general_expectation} for the product $\Zbf_n = \prod_{\ell = 0}^n \Y_\ell$ where $\sequence{\Y}[\ell][\nset]$ are an independent and  for each $\ell, q \in \nset$ there exist $m_\ell \in (0,1)$  and $\sigma_{\ell,q} > 0$ such that \(\norm{\PE[\Y_\ell]}[Q]^2  \leq 1 - m_\ell\) and \(\PE^{1/q}[\norm{\Y_\ell - \PE[\Y_\ell]}[Q]^q] \leq \sigma_{\ell,q}\). We use notations of $\bfA_n, \bfB_n$ from \Cref{th:general_expectation}. Applying independence of $\Zbf_{n-1}$ and $\Y_n$ and \(\PE^{1/q}[\norm{\Y_\ell - \PE[\Y_\ell]}[Q]^q] \leq \sigma_{\ell,q}\) we estimate   
\begin{align}
     \label{eq:bound_MP_1_proof_unbounded}
        \norm{\bfA_n}[p,q] \le  \left(\PE\left[ \normop{\Y_n - \PE [\Y_n]}[Q]^q  \norm{f_Q(\Zbf_{n-1})}[p]^q  \right]\right)^{1/q} \leq\sigma_{n,q} \norm{ f_Q(\Zbf_{n-1})}[p,q].
    \end{align}
The bound for $\norm{\bfB_n}[p,q]^2$ remains the same: $\norm{\bfB_n}[p,q]^2 \leq (1 - m_n) \norm{f_Q(\Zbf_{n-1})}[p,q]^2$.
Combining this inequality with \eqref{eq:bound_MP_1_proof_unbounded} and \eqref{eq:bound_decomp_proof_mat_prod} yields for any $n \in\nsets$, $\norm{f_Q(\Zbf_n)}[p,q]^2 \leq (1 - m_n + (p-1)\sigma_{n,q}^2)  \norm{f_Q(\Zbf_{n-1})}[p,q]^2 \leq \prod_{i=1}^n (1 - m_i + (p-1)\sigma_{i,q}^2)  \norm{f_Q(\Zbf_{0})}[p,q]^2$. 
    The proof is then completed upon using \eqref{eq:submultip}  which implies that $\norm{\Zbf_n}[p,q]= \norm{Q^{-1/2}f_Q(\Zbf_n) Q^{1/2}}[p,q] \leq \sqrt{\qcond} \norm{f_Q(\Zbf_n)}[p,q]$.
Finally, it remains take $\Y_\ell = \Id - \alpha \Am_\ell, \ell \geq 1$, $\Y_0 = \Id$. As $-\bA$ is Hurwitz, applying \Cref{lem:Hurwitzstability} yields $\norm{\PE[\Y_\ell]}[Q]^2 = \norm{\Id - \alpha \bA }[Q]^2 \leq 1 - a \alpha$. Further, since $\norm{\Am_\ell - \bA} \in \SG(\bConst{A}')$ we get by \Cref{lem: pmomSG}
 \[
     \PE^{1/q}[\norm{\Y_\ell - \PE[\Y_\ell]}[Q]^q] =  \alpha \PE^{1/q}[\norm{ \Am_\ell- \bA}[Q]^q] \le   2\alpha \sqrt{\qcond q}  \bConst{A}'  = \alpha b_{Q}'  \sqrt{q} \eqsp. 
 \]
Taking $m_\ell = a \alpha$ and  $\sigma_{\ell,q} = b_Q' \alpha \sqrt{q}$ we get the claim of the proposition.

\end{proof}



\section{Proofs of \Cref{sec:finite-time-high}}
\label{sec:proof_finite_time}
For ease of presentation, we drop in this section the dependence of
$\Jnalpha{n}{0},\Hnalpha{n}{0},\Jnalpha{n}{1},\Hnalpha{n}{1}$ with respect to $\alpha$ and simply write $J_n^{(0)},H_n^{(0)},J_n^{(1)},H_n^{(1)}$, respectively. We denote $\tbfA_n = \Am_n - \bA$.
\subsection{Proof of \Cref{prop:transient_term_bound}}
\label{sec:proof_finite_time_transient}
\label{subsec:proof_transient}
Let $n\in\nsets$, $\alpha \in \ocint{0,\alpha_{p_0,\infty}}$, $u\in\sphere^{d-1}$ and $\delta \in \ooint{0,1}$. Under \Cref{assum:noise}, applying \Cref{corr:concentration_iid} with $p = p_0$ yields
\begin{align*}
\PE^{1/p_0}[|u^\top \ProdBa_{1:n}(\theta_0-\thetalim)|^{p_0}] 
&\leq \PE^{1/p_0}[\norm{\ProdBa_{1:n}}^{p_0}]\norm{\theta_0-\thetalim} \\ 
&\leq \sqrt{\qcond} d^{1/p_0} (1 - a \alpha + (p_0-1) b_Q^2 \alpha^2)^{n/2}\norm{\theta_0-\thetalim}\eqsp.
\end{align*}
Since $\alpha \leq a/(2b_Q^2(p_0-1))$, using $(1-t)^{1/2} \leq 1-t/2$ for $t \in \ccint{0,1}$, we get 
$$
\PE^{1/p_0}[|u^\top \ProdBa_{1:n} \ttheta_0|^{p_0}] \leq \sqrt{\qcond} d^{1/p_0} \norm{\theta_0-\thetalim}(1-a\alpha/4)^{n}\eqsp.
$$
Applying Markov's inequality easily completes the proof.

\subsection{Proof of \texorpdfstring{\Cref{th:HP_bound_J_n_0_rosenthal new}}{Proposition 5}}
\label{subsec:proof_J_n_0_bound}
Let $n\in\nsets$, $\alpha \in \ocint{0,\alpha_{p_0,\infty}}$, $u\in\sphere^{d-1}$ and $\delta \in \ooint{0,1}$. Using \eqref{eq:jn0_main} and applying Rosenthal's inequality \cite[Theorem 4.1]{pinelis_1994}\footnote{Note that the specific universal constants $\bConst{\sf{R},1}=60\rme$ and $\bConst{\sf{R},2}=60$ are not given in the statement, but a close inspection of the proof provide the given estimates.} for sum of centered independent random variables we get for any $p \geq 2$,
$$
\PE[|u^\top J_n^{(0)}|^p] \le  (60 \rme)^p p^{p/2}  \{u^{\transpose}\lineGa_n u \}^{p/2} + \alpha^p  60^p p^p \expe{\max_{\ell = 1, \ldots, n}|u^\top G_{\ell+1:n} \funnoisew_{\ell} |^p} \eqsp.
$$
Applying  \Cref{lem: second term in Ros}, we obtain for any $p \geq 2$,
\begin{equation}
\label{eq:J_n_0_moment_bound}
\PE[|u^\top J_n^{(0)}|^p] \leq  (60 \rme)^p p^{p/2}  \{u^{\transpose} \lineGa_n u \}^{p/2} + (9\{1+\log[1/(a\alpha)]\}\qcond\bConst{\eps}^2 )^{p/2} \alpha^p 60^p p^{3p/2} \eqsp, 
\end{equation}
where the constant $\bConst{\eps}$ is given in \eqref{eq:subgaus_const_epsilon}. Applying Markov's inequality,
we get for any $p \geq 2$, $c_1,c_2 >0$,
\begin{align*}
  &\PP(|u^\top J_n^{(0)}| \geq  c_1 \{u^{\transpose} \lineGa_n u \}^{1/2} + c_2) \\ 
  & \leq \{c_1 \{u^{\transpose} \lineGa_n u \}^{1/2} + c_2\}^{-p} 
    \parentheseDeux{(60 \rme)^p p^{p/2}  \{u^{\transpose} \lineGa_n u \}^{p/2} + (9 \{1+\log[1/(a\alpha)]\}\qcond\bConst{\eps}^2 )^{p/2} \alpha^p 60^p p^{3p/2}} \\
  & \leq  (60 \rme)^p p^{p/2}c_1^{-p} + (9\{1+\log[1/(a\alpha)]\}\qcond\bConst{\eps}^2 )^{p/2} \alpha^p 60^p p^{3p/2} c_2^{-p}\eqsp.
\end{align*}
Taking $p = 3\log{(2/\delta)}$, $c_1 = \sfD_1(\log{(2/\delta)})^{1/2}$ and $c_2 = \alpha \sqrt{1 + \log(1/(a\alpha))} \ConstD_{2} \log^{3/2} (2/\delta)$
yields the statement, where 
\begin{equation}
\label{eq:const_D_1_D_2_def}
\sfD_1 = 60\sqrt{3}\rme^{4/3}, \quad \sfD_2 = 540 \sqrt{3}\rme^{1/3} \qcond^{1/2}\bConst{\eps} \eqsp. 
\end{equation}

\subsection{Proof of \Cref{prop:Sigma_alpha_expansion}}
\label{subsec:proof_var_J_n_0}
\begin{lemma}
\label{lem:ricatti}
Assume that \Cref{assum:noise}-\ref{assum:Hurwitzmatrices} holds. 
Then, for any $\alpha \in (0,\alpha_{\infty}]$, where $\alpha_{\infty}$ is defined in \eqref{eq: kappa_def},
\[
  \norm{\lineGa -  \lineG}[Q] \leq \alpha a^{-1}\|\bA \lineG \bA^\top\|_{Q}   \eqsp,
\]
where $\lineGa$ and $\lineG$ are defined in  \eqref{eq:ricatti} and \eqref{eq:def_lineG} respectively and  $a$ is given in
 \eqref{eq: kappa_def}.
\end{lemma}
\begin{proof}
  Let $\alpha \in (0,\alpha_{\infty}]$.
By definition, \eqref{eq:ricatti} and \eqref{eq:def_lineG} imply
\begin{equation*}
\bA(\lineGa -  \lineG) + (\lineGa - \lineG)\bA^\top - \alpha \bA (\lineGa -  \lineG) \bA^\top = \alpha \bA \lineG \bA^\top\eqsp,
\end{equation*}
which writes 
\begin{equation*}
\lineGa -  \lineG - (\Id - \alpha \bA)(\lineGa -  \lineG)(\Id - \alpha \bA)^\top = \alpha^2 \bA \lineG \bA^\top\eqsp.
\end{equation*}
This implies, by \Cref{lem:Hurwitzstability}, 
\[
\| \lineGa - \lineG\|_{Q} \leq (1-\alpha a)\| \lineG -  \lineG\|_{Q} + \alpha^2 \|\bA \lineG \bA^\top\|_{Q} \eqsp,
\]
Rearranging terms completes the proof.
\end{proof}

\subsection{Proof of \Cref{prop:J_n_1_bound_new}}
\label{subsec:proof_J_n_1_bound}

\Cref{prop:J_n_1_bound_new} is a direct consequence of the following statements.
\begin{Prop}
\label{prop:J_n_1_bound_new_app}
Assume \Cref{assum:noise}. Then, for any $n \in\nset$, $\alpha \in (0,\alpha_{\infty})$, where $\alpha_{p_0,\infty}$ is defined in \eqref{eq:def_alpha_p_infty},  $u \in \sphere^{d-1}$ and $p \geq 2$,
\begin{equation}
\label{eq:moment-bound-J1_app}
\PE[\bigl|u^\top J_{n}^{(1)}\bigr|^p] \leq  \sfD_3^p \alpha^p p^{2p}\eqsp,
\end{equation}
where  $\sfD_{3}$ is given in~\eqref{eq: Jn1 bound}. Moreover, for any
$\delta \in (0,1)$ with probability at least $1 - \delta$,
\begin{equation}
\label{eq:HPB-J1_app}    
\bigl|u^\top J_{n}^{(1)}\bigr| \leq \rme \sfD_3 \alpha \log^2(1/\delta)\eqsp.
\end{equation}
\end{Prop}
\begin{Prop}
\label{prop:H_n_1_bound_new_app}
Assume \Cref{assum:noise} and  let $p_0 \geq 2$. Then, for any $n \in\nset$, $\alpha \in (0,\alpha_{p_0,\infty})$, where $\alpha_{p_0,\infty}$ is defined in \eqref{eq:def_alpha_p_infty},  $u \in \sphere^{d-1}$,
\begin{equation}
\label{eq:moment-bound-h1_app}
\PE[\bigl| u^\top H_{n}^{(1)} \bigr|^{p_0}] \leq \ConstD_4^{p_0} \alpha^{p_0} {p_0}^{2p_0}\eqsp,
\end{equation}
where $\ConstD_4$ is given in \eqref{eq:moment_bound_H_n_1}. Moreover, for any
$\delta \in (0,1)$ with probability at least $1 - \delta$,
\begin{equation}
\label{eq:HPB-H1_app}    
 \bigl| u^\top H_{n}^{(1)} \bigr| \leq \ConstD_4 \alpha p_0^2 \delta^{-1/p_0}\eqsp.
\end{equation}
\end{Prop}

\begin{proof}[Proof of \Cref{prop:J_n_1_bound_new_app}]
First, we note that \eqref{eq:expansion_recur_gen} implies 
\begin{equation*}
J_{n}^{(1)} = \alpha^2 \sum_{\ell=1}^{n-1} S_{\ell+1:n} \funnoisew_{\ell}, ~~\text{with} ~~ S_{\ell+1:n} = \sum_{k=\ell+1}^{n} (\Id - \alpha \bA)^{n-k-1} \tbfA_{k} (\Id - \alpha \bA)^{k-1 - \ell} \eqsp.
\end{equation*}
It is easy to check that the sequence $( \alpha^2 S_{\ell+1:n} \funnoisew_{\ell}, \mathfrak F_{\ell+1:n})_{\ell=1}^{n-1}$ is a martingale-difference, where $\mathfrak F_{\ell+1:n} = \sigma\left((\Am_{j},\bm_{j})_{j \in \{\ell+1,\dots,n\}}\right)$. We may use Burkholders's inequality \cite[Theorem 2.10]{hallheydebook} to get
\begin{equation}
\label{eq:pinelis_J_n_1}
\PE[ \bigl|u^\top J_{n}^{(1)}\bigr|^p ] \leq (36 p)^p \alpha^{2p} \PE \Biggl [ \Bigl(\sum_{\ell=1}^{n-1} (u^\top S_{\ell+1:n} \funnoisew_{\ell})^2 \Bigr)^{p/2}\Biggr].
\end{equation}
Using the Minkowski inequality,
\begin{align}
\label{eq:pinelis_J_n_1 2}
\PE[ \bigl|u^\top J_{n}^{(1)}\bigr|^p ] \leq (36 p)^p \alpha^{2p}  \Bigl( \sum_{\ell=1}^{n-1}  \PE^{2/p}[(u^\top S_{\ell+1:n} \funnoisew_{\ell})^p] \Bigr)^{p/2} .
\end{align}
Denote $V_{\ell+1}^\top = u^\top S_{\ell+1:n}$.
Note that by Assumption \Cref{assum:noise}-\ref{assum:Abounded} and Lemma~\ref{lem:Hurwitzstability}, $\norm{(\Id - \alpha \bA)^{n-k-1} \tbfA_{k} (\Id - \alpha \bA)^{k-1 - \ell}} \leq \qcond \bConst{A}(1-\alpha a)^{(n-\ell-2)/2}$. Applying \cite[Theorem~3]{Pinelis1992}\footnote{with $\mathcal{X} = \rset^d$ equipped with the Euclidean norm $\norm{\cdot}$. Note that $\norm{x}, x \in \rset^d$ is twice Gateaux differentiable and $\mathcal{X} \in D(A_1,A_2)$ with $A_1 = A_2 = 1$.}, we get for any $t \geq 0$
\begin{align*}
\PP\bigl(\norm{V_{\ell+1}} \geq t\bigr) \leq 2\exp\left\{-t^2/\left(2\qcond^2\bConst{A}^2(n-\ell)(1-\alpha a)^{n-\ell-2}\right)\right\}\eqsp.
\end{align*}
Using \Cref{lem: pmomSG},
\begin{equation}
\label{eq:l_p_bound_V_ell}
\PE^{2/p}[\norm{V_{\ell+1}}^p] \leq 2\sqrt{2}\bConst{A}^2\qcond^2(n-\ell)(1-\alpha a)^{(n-\ell-2)}p.
\end{equation} 
Since $S_{\ell+1:n}$ and $\funnoisew_{\ell}$ are independent, 
$$
\PE[|u^\top S_{\ell+1:n}\funnoisew_{\ell}|^p] \leq \PE [\normop{V_{\ell+1}}^p] \sup_{u \in \sphere^{d-1}} \PE[|u^\top\funnoisew_{\ell}|^p ]\,.
$$
Assumption \Cref{assum:noise}-\ref{assum:b_subexp}, Lemma \ref{lem: pmomSG} and \Cref{lem: second term in Ros} imply, that for any $u \in \sphere^{d-1}$,
\begin{align*}
\PE[|u^\top\funnoisew_{\ell}|^p ] \leq p^{p/2} \bConst{\eps}^p (2 \sqrt 2)^{p/2}  
\eqsp.
\end{align*}
Combining this bound with \eqref{eq:l_p_bound_V_ell},
\begin{equation}
\label{eq: Jn1 bound}
 \PE[|u^\top S_{\ell+1:n}\funnoisew_{\ell}|^p]
 \leq p^{p} (2\sqrt{2})^{p}   \bConst{A}^p\qcond^p \bConst{\eps}^p (n-\ell)^{p/2}(1-\alpha a)^{(n-\ell-2)p/2} \eqsp.
\end{equation}
This inequality and \eqref{eq:pinelis_J_n_1 2} imply 
\begin{equation}
\label{eq:Jn1_moment_bound}
\begin{split}
   \PE[ \bigl|u^\top J_{n}^{(1)}\bigr|^p ] &\leq p^{2p} \alpha^{2p}  (72\sqrt{2})^{p}   \bConst{A}^p\qcond^p \bConst{\eps}^p \Bigl( \sum_{\ell=1}^{n-1} (n-\ell) (1-\alpha a)^{(n-\ell-2)} \Bigr)^{p/2} \\
   & \le \alpha^p \sfD_3^p p^{2p}, \quad \text{where} \quad \sfD_3 =  (72\sqrt{2}) \bConst{A}\qcond \bConst{\eps} a^{-1} (1 - a \alpha_\infty)^{-1}.
\end{split}
\end{equation}
Now the equation~\eqref{eq:HPB-J1} follows from Markov's inequality. Namely, for any $c_1 > 0$ it holds
\begin{equation*}
\PP\left(\bigl|u^\top J_{n}^{(1)}\bigr| \geq c_1 \alpha \sfD_3 \right) \leq \frac{\alpha^{p}\sfD_3^p p^{2p}}{c_1^p \alpha^p\sfD_3^p} = c_1^{-p}p^{2p}\,.
\end{equation*}
Taking $p = \log{(1/\delta)}$ and $c_1 = \rme \log^2{(1/\delta)}$, we obtain \eqref{eq:HPB-J1}.
\end{proof}

\begin{proof}[Proof of \Cref{prop:H_n_1_bound_new_app}]
With the decomposition \eqref{eq:expansion_recur_gen}, we represent
\begin{equation*}
u^\top H_{n}^{(1)} = -\alpha \sum_{\ell=1}^{n} u^\top\ProdBa_{\ell+1:n} \tbfA_{\ell} J_{\ell-1}^{(1)}.
\end{equation*}
Using Minkowski's inequality,
$$
\PE^{1/p}[\bigl| u^\top H_{n}^{(1)} \bigr|^p] \leq \alpha \sum_{\ell=1}^{n}\PE^{1/p}[\bigl| u^\top \ProdBa_{\ell+1:n} \tbfA_{\ell} J_{\ell-1}^{(1)}\bigr|^p]\eqsp.
$$
Now, using the independence of $\ProdBa_{\ell+1:n}$, $\tbfA_{\ell}$, $J_{\ell-1}^{(1)}$, and \Cref{assum:Abounded}, 
\begin{equation}
\label{eq:Gamma_ell_prod_bound}
\begin{split}
\PE^{1/p}[\bigl| u^\top \ProdBa_{\ell+1:n} \tbfA_{\ell} J_{\ell-1}^{(1)}\bigr|^p] 
&\leq \PE^{1/p}\bigl[\norm{u^\top \ProdBa_{\ell+1:n} \tbfA_{\ell}}\bigr]\sup_{v \in \sphere^{d-1}}\PE^{1/p}\bigl[|v^\top J_{\ell-1}^{(1)}|\bigr] \\
&\leq 2\bConst{A}\PE^{1/p}[\normop{\ProdBa_{\ell+1:n}}^{p}]\sup_{v \in \sphere^{d-1}}\PE^{1/p}\bigl[|v^\top J_{\ell-1}^{(1)}|\bigr]\eqsp.
\end{split}
\end{equation}
Hence, applying \Cref{corr:concentration_iid} to $\PE^{1/p}[\normop{\ProdBa_{\ell+1:n}}^{p}]$, and \eqref{eq:Jn1_moment_bound} to $\sup_{ v \in \sphere^{d-1}}\PE^{1/p}[\bigl|v^\top J_{\ell-1}^{(1)} \bigr|^{p}]$,
$$
\PE^{1/p}[\bigl| u^\top H_{n}^{(1)} \bigr|^p] \leq 2\bConst{A} \sqrt{\qcond} d^{1/p} \ConstD_{3} \alpha^2  p^2 \sum_{\ell = 1}^{n}  \bigl(1 - a \alpha + (p-1) b_Q^2 \alpha^2\bigr)^{(n-\ell)/2}\eqsp.
$$
Since $p_0 -1 \leq a/(2b_Q^2 \alpha)$, from the previous estimate it follows
\begin{align*}
\PE^{1/p_0}[\bigl| u^\top H_{n}^{(1)} \bigr|^{p_0}] 
&\leq 2\bConst{A} \sqrt{\qcond} d^{1/p_0} \ConstD_{3} \alpha^2  p_0^2 \sum_{\ell = 1}^{n}(1-\alpha a)^{(n-\ell)/2}\\
& \leq 4\bConst{A} \sqrt{\qcond} d^{1/p_0} \ConstD_{3} \alpha  p_0^2/a\eqsp.
\end{align*}
Hence, 
\begin{equation}
\label{eq:moment_bound_H_n_1}
\PE[\bigl| u^\top H_{n}^{(1)} \bigr|^{p_0}] \leq \ConstD_4^{p_0} \alpha^{p_0} {p_0}^{2p_0}\eqsp, \text{ where } \ConstD_4 = 4\bConst{A} \sqrt{\qcond} d^{1/p_0} \ConstD_{3}/a\eqsp.
\end{equation}
Using Markov's inequality, we get with probability at least $1 - \delta$,
$$
\bigl|u^\top H_{n}^{(1)}\bigr| \leq \ConstD_4 \alpha p_0^2/\delta^{1/p_0}\eqsp.  
$$
\end{proof}

\subsection{Proof of \Cref{prop:H_n_1_bound_contractive}}
\label{subsec:proof_H_n_1_bound_A2}
The proof is along the same lines as the proof of $H_n^{(1)}$ in \Cref{subsec:proof_J_n_1_bound}, with the better bound for $\PE^{1/p}[\normop{\ProdBa_{\ell+1:n}}^{p}]$. For reader's convenience, we provide the proof below. Starting with equation \eqref{eq:Gamma_ell_prod_bound}, we note that under \Cref{ass:contraction},
\[
\PE^{1/p}[\normop{\ProdBa_{\ell+1:n}}^{p}] \leq \sqrt{\qcondtild} \bigl(1 - \alpha \ta \bigr)^{n-\ell}\eqsp.
\]
We also apply \eqref{eq:Jn1_moment_bound} to $\sup_{ v \in \sphere^{d-1}}\PE^{1/p}[\bigl|v^\top J_{\ell-1}^{(1)} \bigr|^{p}]$. Then
$$
\PE^{1/p}[\bigl| u^\top H_{n}^{(1)} \bigr|^p] \leq 2 \sqrt{\qcondtild}\bConst{A} \ConstD_{3} \alpha^2  p^2 \sum_{\ell = 1}^{n}\bigl(1 - \alpha \ta \bigr)^{n-\ell} \leq \ConstD_5 \alpha p^2\eqsp,
$$
where we have defined 
\begin{equation}
\label{eq:const_D_5_definition}
\ConstD_5 = 2 \sqrt{\qcondtild}\bConst{A} \ConstD_{3}/\ta\eqsp.
\end{equation}
Now the equation~\eqref{eq:HPB-H1-contractive} follows from Markov's inequality.

\section{Concentration results for sub-Gaussian random variables}

\begin{lemma}
\label{lem: pmomSG}
Random variable $X \in \SG(\sigma^2)$ for some $\sigma > 0$ if and only if for all $t \geq 0$ the condition $\PP(|X| \geq t) \leq 2 \exp\{-t^2/(2\sigma^2)\}$ holds. In addition, in such a case, for any $p \geq 2$, we have
$$
\PE[|X|^p] \leq \sqrt{2} \rme (2/\rme)^{p/2} p^{p/2} \sigma^p.
$$
\end{lemma}
\begin{proof}
The first statement is well-known, see for example \cite[Theorem~2.1]{blm:2013}.
We now show the second statement. 
By the Fubini theorem,  $\PE[|X|^p] = p\int_{0}^{+\infty}u^{p-1}\PP(|X| > u)\,\rmd u$, we get
\begin{equation*}
\PE[|X|^p] = 2p \int_0^\infty u^{p-1} \rme^{-u^2/(2\sigma^2)} \, \rmd u  = p 2^{p/2} \sigma^p \Gamma(p/2),
\end{equation*}
using the change of variable $t = u^2/(2\sigma^2)$. Now, using an upper bound $\Gamma(p/2) \le (p/2)^{(p-1)/2} \rme^{1 - p/2}$ (see e.g. \cite[Theorem~2]{guo:bounds_for_gamma_function}), and $p^{1/2} \leq 2^{p/2}$, we finally get
\begin{equation*}
\PE[|X|^p] \leq \sqrt{2} \rme (2/\rme)^{p/2} p^{p/2} \sigma^p  \eqsp.
\end{equation*}
\end{proof}

\begin{lemma}
\label{lem:max_subgaus_moment}
Let $\sequence{X}[\ell][\nset]$ be a sequence of random variables such that $X_\ell \in \SG(\sigma^2)$ for any $ \ell \in \nset$ and some $\sigma^2 >0$. Then for any $p \geq 2$,
\begin{equation*}
\PE\left[\max_{\ell = 1,\dots,n} \left(|X_\ell|/\sqrt{1 + \log \ell}\right)^p\right] \leq 3^p \sigma^p p^{p/2}\eqsp.
\end{equation*}
\end{lemma}
\begin{proof}
  Set $a_k = (1+\log{k})^{1/2}$ for $k \in\nsets$. Using the Fubini's
  theorem,
  $\PE[|\xi|^p] = p\int_{0}^{+\infty}u^{p-1}\PP(|\xi| > u)\rmd u$, the
  union bound, and \Cref{lem: pmomSG}, we get 
\begin{align*}
&\PE\biggl[\max_{k = 1,\dots,n}\{|X_k|^p/a_k^p \}\biggr] 
= p\int_{0}^{+\infty}u^{p-1}\PP\parenthese{\max_{k = 1,\dots,n}|X_k| \geq u a_k}\rmd u \\
& \qquad \qquad \qquad \leq 2^p\sigma^p + p\int_{2\sigma}^{+\infty}u^{p-1}\PP\parenthese{\max_{k = 1,\dots,n}|X_k| \geq u a_k}\,\rmd u \\
&\qquad \qquad \qquad \leq 2^p\sigma^p + p\int_{2\sigma}^{+\infty}u^{p-1}\sum_{k=1}^{n}\PP\bigl(|X_k| \geq u a_k\bigr)\,\rmd u \\
& \qquad \qquad \qquad\leq 2^p\sigma^p + 2p\int_{2\sigma}^{+\infty}u^{p-1}\sum_{k=1}^{n}\exp\bigl\{-u^2a_k^2/(2\sigma^2)\bigr\}\,\rmd u \\
&\qquad \qquad \qquad\leq 2^p\sigma^p + 2p\sigma^{p}\int_{2}^{+\infty}y^{p-1}\exp\bigl\{-y^2/2\bigr\}\bigl(\sum_{k=1}^{n}k^{-y^2/2}\bigr)\,\rmd y\\
&\qquad \qquad \qquad \leq 2^p\sigma^p + \frac{\uppi^2 p \sigma^{p}}{3}\int_{2}^{+\infty}y^{p-1}\exp\bigl\{-y^2/2\bigr\}\,\rmd y\leq 2^p\sigma^p + \frac{\uppi^2 p \sigma^{p}2^{p/2-1}}{3}\Gamma(p/2)\,.
\end{align*}
Using $\Gamma(p/2) \le (p/2)^{(p-1)/2} \rme^{1 - p/2}$ (see \cite[Theorem~2]{guo:bounds_for_gamma_function}), we get
$$
\PE\biggl[\max_{k = 1,\dots,n}\{|X_k|^p/a_k^p \}\biggr]  \le 2^p\sigma^p + \uppi^2 \sigma^p p^{(p+1)/2} \rme^{1 - p/2}/(3\sqrt{2}) \eqsp.
$$
Since $\sqrt{p}\rme^{-p/2} \leq \rme^{-1/2}$ and $2^p \leq 4p^{p/2}$, 
$$
\PE\biggl[\max_{k = 1,\dots,n}\{|X_k|^p/a_k^p \}\biggr]  \leq \sigma^p p^{p/2}\bigl(4 + \uppi^2\rme^{1/2}/(3\sqrt{2})\bigr) < 9\sigma^p p^{p/2}\eqsp.
$$
\end{proof}

\begin{lemma}
  \label{lem: second term in Ros}
  Assume \Cref{assum:noise}. Then, for any $n \in\nsets$ and $v \in \sphere^{d-1}$, $v^{\transpose} \funnoisew_n$ defined by \eqref{eq:LSA-recursion-main} is a sub-Gaussian random variable with parameter
\begin{equation}
\label{eq:subgaus_const_epsilon}
\bConst{\eps}^2 = 2\bConst{b}^2 + 8\bConst{A}^2\normop{\thetalim}^2\eqsp. 
\end{equation}
In addition, for any $n \in \nsets$, $p \geq 2$, $u \in\sphere^{d-1}$ and $\alpha \in \ooint{0,\alpha_{\infty}}$, it holds
\begin{equation*}
\PE\left[\max_{\ell= 1,\dots, n}|u^\top \Ga_{\ell+1:n} \funnoisew_{\ell}|^p\right] \leq \bigl(9 \qcond p \bConst{\eps}^2\{1 + \log[1/(a\alpha)]\}\bigr)^{p/2}\, \eqsp,
\end{equation*}
where  $\alpha_{\infty}$, $a$ and $\qcond$  are defined in \eqref{eq: kappa_def}.
\end{lemma}
\begin{proof}
First we prove \eqref{eq:subgaus_const_epsilon}. Using the representation \eqref{eq:LSA-recursion-main}, for any $\lambda \in \rset$,
\begin{align*}
\PE\bigl[\exp\left\{\lambda v^\top \funnoisew_n\right\}\bigr] &\leq \PE\bigl[\exp\left\{\lambda v^\top (\bm_{n} - \barb - \{ \Am_{n} - \bA \} \thetas)\right\}\bigr] \\
&\leq \PE^{1/2}\bigl[\exp\left\{2\lambda v^\top (\bm_{n} - \barb)\right\}\bigr]\PE^{1/2}\bigl[\exp\left\{2\lambda v^\top(\bA - \Am_{n})\thetas\right\}\bigr]\eqsp.
\end{align*}
Note that \Cref{assum:noise}-\ref{assum:Abounded} implies $\left|v^\top (\bA - \Am_{n})\thetas\right| \leq 2\bConst{A}\norm{\thetas}$. Hence, using the Hoeffding inequality, $v^\top (\bA - \Am_{n})\thetas \in \SG(4\bConst{A}^2\norm{\thetas}^2)$. Combining this result with \Cref{assum:noise}-\ref{assum:b_subexp},
\begin{align*}
\PE\bigl[\exp\left\{\lambda v^\top \funnoisew_n\right\}\bigr] \leq \exp\left\{\lambda^2 \bConst{b}^2\right\} \exp\left\{4\lambda^2 \bConst{A}^2\norm{\thetas}^2\right\}\eqsp,   
\end{align*}
yielding the first statement of the lemma.
\par
To prove the second part, let us denote $v_{\ell} = (\Id - \alpha \bA)^{n - \ell}u/\norm{(\Id- \alpha \bA)^{n-\ell}u} \in \sphere^{d-1}$. Using \Cref{lem:Hurwitzstability},
\begin{align*}
&\PE[\max_{\ell= 1,\dots, n}|u^\top \Ga_{\ell+1:n} \funnoisew_{\ell}|^p] 
= \PE[\max_{\ell= 1,\dots, n}|v_{\ell}^\top\funnoisew_{\ell}|^p\norm{\Ga_{\ell+1:n}u}^p] \\
&\leq  \qcond^{p/2} \PE[\max_{\ell= 1,\dots, n}|v_{\ell}^\top\funnoisew_{\ell}|^p (1-\alpha a)^{p(n-\ell)/2}]\\
&\leq \qcond^{p/2} \PE\biggl[\max_{\ell= 1,\dots, n}\frac{|v_{\ell}^\top\funnoisew_{\ell}|^p}{(1+ \log{(n-\ell+1)})^{p/2}}\biggr]\biggl\{\max_{x> 0} (1+ \log(x+1))\rme^{-a\alpha x})\biggr\}^{p/2} \\
  &\leq \qcond^{p/2} (9\bConst{\eps}^2 p)^{p/2}\biggl\{\max_{x> 0} [(1+ \log(x+1))\rme^{-a\alpha x}]\biggr\}^{p/2}  \eqsp,
\end{align*}
where in the last inequality we used \Cref{lem:max_subgaus_moment}.
Set $f(x)=(1+ \log(x+1))\rme^{-c x}$ with $c=a\alpha \leq 1$ over $x>0$. First, note that $f'(x)=\rme^{-c x}(1/(1+x) - c - c\log(x+1)) < 0$ for all $x>1/c - 1$, and thus the maximum is attained for $x\in[0,1/c - 1]$. Moreover, for any $ x\leq 1/c - 1$, we have $f(x)\leq 1+\log(1+x) \leq 1+\log(1/c)$, leading to the desired result.
\end{proof}


\section{Proof of \Cref{sec:optim-deriv-bounds}}

\subsection{Proof of \Cref{theo:convergence_mc_wasserstein}}
\label{sec:proof-crefth_mc_wasser}
Let $\alpha\in \ooint{0,\alpha_{2,\infty}}$ and  $\lambda_1,\lambda_2 \in \Pens_2(\rset^d)$. By \cite[Theorem
  4.1]{villani:2009}, there exists a couple of random variables
  $\theta^{(1)}_0,\theta^{(2)}_0$ such that
  $W_2^2(\lambda_1,\lambda_2) =
  \PE[\normLigne{\theta^{(1)}_0-\theta^{(2)}_0}^2]$ independent of  $\sequenceDouble{\Am}{\bm}[n][\nsets]$. We introduce then a synchronous coupling between $\lambda_1 R_{\alpha}^n$ and $\lambda_2 R_{\alpha}^n$ as follows. Let
  $\sequenceDouble{\theta^{(1)}}{\theta^{(2)}}[n][\nset]$  starting from
  $\theta^{(1)}_0$ and $\theta^{(2)}_0$ respectively and for all $n \geq 0$,
  \begin{equation}
    \label{eq:def_coupling}
    \begin{aligned}
  \theta_{n+1}^{(1)}&=  (\Id-\alpha \Am_{n+1})\theta_n^{(1)} +\alpha \bm_{n+1} \\
  \theta_{n+1}^{(2)} &= (\Id-\alpha \Am_{n+1})\theta_n^{(   2)} + \alpha \bm_{n+1}\eqsp.
\end{aligned}
\end{equation}
Since for all $n \geq 0$, the distribution of
$(\theta_{n}^{(1)},\theta_{n}^{(2)})$ belongs to
$\Pi(\lambda_1 R_{\alpha}^{n},\lambda_2 R_{\alpha}^{n})$,
 by definition of the Wasserstein distance we get for any $n \in \nset$,
 \begin{multline}
     \label{eq:coupling_2}
     W_2( \lambda_1 R_{\alpha}^n, \lambda_2  R_{\alpha}^n) \le \PE^{1/2}[ \norm{\theta_n^{(1)} - \theta_n^{(2)}}^2] = \PE^{1/2}[ \norm{\Gamma_{1:n}[\theta_0^{(1)} - \theta_0^{(2)}]}^2] \\ \leq \ConstD_{2} (1 - a \alpha/2)^{n/2}W_2(\lambda_1,\lambda_2) \eqsp,
   \end{multline}
where we have used \Cref{cor:norm_Gamma_m_n} for the last inequality. 
  By \cite[Theorem
  6.16]{villani:2009}, the space $\mathcal{P}_2(\rset^d)$ endowed with $W_2$ is a Polish space. Then,  $(\lambda_1 R_{\alpha}^n)_{n\geq 0}$ is a Cauchy sequence and converges to a limit $\pi_\alpha^{\lambda_1} \in \mathcal{P}_2(\rset^d)$, $  \lim_{n \to \plusinfty} W_2(\lambda_1 R_{\alpha}^n,  \pi_\gamma^{\lambda_1})=  0$.
We show that the limit $\pi_\alpha^{\lambda_1}$ does not depend on
$\lambda_1$. Assume that there exists $\pi_{\alpha}^{\lambda_2}$ such
that
$ \lim_{k \to \plusinfty} W_2(\lambda_2 R_{\alpha}^n,
\pi_\alpha^{\lambda_2})= 0 $. By the triangle inequality
  $$W_2( \pi_\alpha^{\lambda_1}, \pi_\alpha^{\lambda_2} ) \le W_2(
  \pi_\alpha^{\lambda_1},\lambda_1 R_{\alpha}^n )+ W_2( \lambda_1
  R_{\alpha}^n, \lambda_2 R_{\alpha}^n)+ W_2(\pi_\alpha^{\lambda_2}, \lambda_2
  R_{\alpha}^n) \eqsp.$$ Thus by \eqref{eq:coupling_2}, taking the limits as $n\to \plusinfty$, we
  get $W_2( \pi_\alpha^{\lambda_1}, \pi_\alpha^{\lambda_2} )=0$ and
  $\pi_\alpha^{\lambda_1}=\pi_\alpha^{\lambda_2}$. The limit is thus
  the same for all initial distributions and is denoted by
  $\pi_{\alpha}$.
    Moreover, $\pi_\alpha$ is invariant for $R_{\alpha}$. Indeed for all $k \in \nset^{*}$, $W_2(\pi_{\alpha} R_{\alpha} , \pi_{\alpha}) \leq    W_2(\pi_{\alpha} R_{\alpha} , \pi_{\alpha} R_{\alpha}^n)+  W_2(\pi_{\alpha} R_{\alpha}^n,\pi_{\alpha})$,
    Using \eqref{eq:coupling_2} again, we get taking $n \to \plusinfty$, $   W_2(\pi_{\alpha} R_{\alpha} , \pi_{\alpha})=0$ and $\pi_{\alpha} R_{\alpha}  = \pi_{\alpha}$. The fact that $\pi_{\alpha}$ is the unique stationary distribution is straightforward by contradiction and using \eqref{eq:coupling_2}. \eqref{eq:conv_wasser_pi_alpha} is a simple consequence of \eqref{eq:coupling_2} taking $\lambda_2=\pi_{\alpha}$.

    It remains to show that $ \thetalima$ is well-defined and has distribution $\pi_{\alpha}$. Since $\sequenceDouble{\Am}{\bm}[k][\nsetm]$ is \iid, $\sum_{n\leq -1}\PE^{1/2}[\norm{\theta_{n}-\theta_{n+1}}^2] = \sum_{n\leq -1}   \PE^{1/2}[\norm{\Gamma_{n:0} \bm_{n-1}}^2] = \sum_{n\leq -1}   \PE^{1/2}[\norm{\Gamma_{n:0}}^2]\PE^{1/2}[\norm{ \bm_{n-1}}^2]$ and therefore \Cref{assum:noise}-\ref{assum:b_subexp} combined with \Cref{cor:norm_Gamma_m_n} ensures that this series is finite and therefore $(\theta_n)_{n \in \nsetm}$ defined in \eqref{eq:def_theta_infty_alpha} is a Cauchy sequence almost surely and in $\rml^2$ which ensures its convergence.
  Finally, assume now that $\sequenceDouble{\Am}{\bm}[k][\nsetm]$ is independent of $\sequenceDouble{\Am}{\bm}[k][\nsets]$. To conclude it is then sufficient to note that if $\theta_0 = \thetalima$, then $\theta_1$ has the same distribution as $\thetalima$ by definition of the recursion \eqref{eq:lsa}.

  \subsection{Proof of \Cref{propo:clt_J_n_0}}
\label{sec:proof-crefpr_clt_j}
  Consider a sequence $\sequence{\alpha}[n][\nset]$ converging to $0$ such that for any $n\in\nset$ $ \alpha_n \in \ocint{0,\alpha_{2,\infty}}$, and let $u \in \sphere^{d-1}$. For ease of notation, we simply denote $G_{k:0}^{(n)} = G_{k:0}^{(\alpha_n)}$.    Note that
  \begin{equation*}
    \alpha^{-1/2}_n u^{\transpose} J_{\infty}^{(\alpha_n,\leftarrow)} = \sum_{k=-\infty}^{1} \Delta M_{n,k} \eqsp, \qquad \Delta M_{n,k} =   \alpha_n^{1/2} u^{\transpose} G^{(n)}_{k:0} \funnoisew_{k-1} \eqsp. 
  \end{equation*}
  By \cite[Theorem 3.6]{hallheydebook}, it is sufficient to show that
  \begin{align}
    \label{proof_tcl_j_1}
    &   \sup_{k \leq 1}  \Delta M_{n,k} \convproba{n \to \plusinfty}  0 \\
        \label{proof_tcl_j_2}
    &   \sum_{k\leq 1} \Delta M_{n,k}^2 \convproba{n\to \plusinfty} u^{\transpose} \lineG u \\
        \label{proof_tcl_j_3}
    & \sup_{n \in \nset} \PE[ \sup_{k \leq 1} \Delta M_{n,k}^2] < \plusinfty \eqsp. 
  \end{align}
  First, by Markov inequality and \Cref{lem:Hurwitzstability} and \Cref{assum:noise}-\ref{assum:b_subexp}, we have
  for any $\eta >0$ that
  \begin{align*}
   & \PP(     \sup_{k \leq 1}  \Delta M_{n,k} \geq \eta) \leq \eta^{-4} \PE[ \sup_{k \leq 1}  \Delta M_{n,k}^4]  \leq \eta^{-4} \alpha_n^{2} \sum_{k \leq 1} \PE[\norm{G^{(n)}_{k:0}}^4 \norm{\funnoisew_{k-1}}^4] \\
    & \leq \eta^{-4} \alpha_n^{2}\PE[\norm{\funnoisew_{0}}^4] \qcond^2\sum_{k \leq 1}(1-a \alpha_n)^{-2(k-1)} \leq \eta^{-4} \alpha_n^{2}\PE[\norm{\funnoisew_{0}}^4] \qcond^2(1-(1-a\alpha_n)^2)^{-1} \eqsp,
  \end{align*}
  which shows that \eqref{proof_tcl_j_1} holds.

Denote by  $\lineG_n$ the unique solution of the Ricatti equation \eqref{eq:ricatti} with $\alpha \leftarrow \alpha_n$. We get by \Cref{lem:ricatti} that there exists $C \geq 0$ such that for any $n \in \nset$,
  \begin{equation*}
    \norm{\lineG - \lineG_n} \leq C \alpha_n \eqsp.
  \end{equation*}
Therefore, we obtain that 
  \begin{align*}
\abs{    \sum_{k \leq 1} \Delta M_{n,k}^2 - u^{\transpose} \lineG u} \leq \alpha_n \abs{\sum_{k \leq 1} ( (G_{k:0}^{(n)})^{\transpose} u )^{\transpose} [\varepsilon_{k-1}\varepsilon_{k-1}^{\transpose} - \lineG_{\varepsilon}] (G_{k:0}^{(n)})^{\transpose} u } + C \alpha_n  \eqsp. 
  \end{align*}
Then, to establish \eqref{proof_tcl_j_2}, it remains to show that
\begin{equation}
  \label{proof_tcl_j_2_2}
    \alpha_n \abs{\sum_{k \leq 1} ( (G_{k:0}^{(n)})^{\transpose} u )^{\transpose} [\varepsilon_{k-1}\varepsilon_{k-1}^{\transpose}- \lineG_{\varepsilon}] (G_{k:0}^{(n)})^{\transpose} u }\convproba{n \to \plusinfty} 0 \eqsp. 
  \end{equation}
  This follows from \Cref{assum:noise}-\ref{assum:Abounded}-\ref{assum:b_subexp} and  \Cref{lem:Hurwitzstability} which shows that
  \begin{align*}
    &   \txts \PE\parentheseDeuxLigne{    \abs{\sum_{k \leq 1} ( (G_{k:0}^{(n)})^{\transpose} u )^{\transpose} [\varepsilon_{k-1}\varepsilon_{k-1}^{\transpose}- \lineG_{\varepsilon}] (G_{k:0}^{(n)})^{\transpose} u }^2} \\
    & \qquad = \sum_{k\leq 1}  \expe{ \abs{( (G_{k:0}^{(n)})^{\transpose} u )^{\transpose} [\varepsilon_{k-1}\varepsilon_{k-1}^{\transpose}- \lineG_{\varepsilon}] (G_{k:0}^{(n)})^{\transpose} u }^2} \\
    & \qquad \leq  \sum_{k \leq 1} \norm{G_{k:0}^{(n)}}^4 \PE[\norm{\varepsilon_{k-1}\varepsilon_{k-1}^{\transpose}- \lineG_{\varepsilon}}^2]   \leq \PE[\norm{\varepsilon_{0}\varepsilon_{0}^{\transpose}- \lineG_{\varepsilon}}^2] \qcond^2(1-(1-a\alpha_n)^2)^{-1} \eqsp. 
  \end{align*}
  Therefore,
  \begin{equation*}
\lim_{n\to \plusinfty} \alpha_n^2    \PE\parentheseDeuxLigne{    \abs{\sum_{k \leq 1} ( (G_{k:0}^{(n)})^{\transpose} u )^{\transpose} [\varepsilon_{k-1}\varepsilon_{k-1}^{\transpose}- \lineG_{\varepsilon}] (G_{k:0}^{(n)})^{\transpose} u }^2} =0 \eqsp,
  \end{equation*}
  which completes the proof of  \eqref{proof_tcl_j_2_2}.

  Finally, we show \eqref{proof_tcl_j_3} which follows from \Cref{assum:noise}-\ref{assum:Abounded}-\ref{assum:b_subexp} and  \Cref{lem:Hurwitzstability}, 
  \begin{align*}
    \PE[ \sup_{k \leq 1} \Delta M_{n,k}^2] \leq \sum_{k \leq 1} \PE[\Delta M_{n,k}^2] & = \alpha_n \sum_{k \leq 1} ( (G_{k:0}^{(n)})^{\transpose} u )^{\transpose} \PE[\varepsilon_{k-1}\varepsilon_{k-1}^{\transpose}](G_{k:0}^{(n)})^{\transpose} u  \\
    & \leq \alpha_n \PE[\varepsilon_{0}\varepsilon_{0}^{\transpose}] \sum_{k \leq 1} \norm{G_{k:0}^{(n)}}^2 \leq  \PE[\varepsilon_{0}\varepsilon_{0}^{\transpose}] \qcond/a \eqsp. 
  \end{align*}


\end{document}